%% file: main_arxiv.tex
\documentclass{article}

\PassOptionsToPackage{numbers, compress}{natbib}

\usepackage[preprint]{neurips_2025}\makeatletter\renewcommand{\@noticestring}{}\makeatother

\usepackage[utf8]{inputenc} 
\usepackage[T1]{fontenc}    
\usepackage{hyperref}       
\usepackage{url}            
\usepackage{booktabs}       
\usepackage{amsfonts}       
\usepackage{nicefrac}       
\usepackage{microtype}      
\usepackage{xcolor}         
\usepackage{amsmath}
\usepackage{multirow}
\usepackage{tikz}
\usepackage{color}
\usepackage{pifont}
\usepackage{amssymb}
\usepackage{wrapfig}
\usepackage{caption} 
\usepackage{wrapfig}
\usepackage{enumitem}
\usepackage{pgfplots}
\usepackage{colortbl}
\usepackage[dvipsnames]{xcolor}
\definecolor{green_im}{rgb}{0.1, 0.55, 0.3}

\newcommand{\rise}[1]{\textcolor{green_im}{\tiny{$\uparrow$#1}}}
\usepackage{xspace}
\usepackage{amssymb}
\usepackage{pifont}
\usepackage{wrapfig}
\usepackage{subcaption}
\usepackage{graphicx}
\usepackage{multicol}
\usepackage{colortbl}
\newcommand{\eg}{\textit{e.g.}}
\newcommand{\ie}{\textit{i.e.}}

\usepackage{enumitem}
\usepackage{tabulary}
\usepackage{tabulary,multirow,overpic,xcolor}
\newcolumntype{x}[1]{>{\centering\arraybackslash}p{#1pt}}

\newlength\savewidth

\title{Revisiting Natural Language Prompts from Large Language Models for Open-Vocabulary Video Recognition}

\author{%
  Haichen He$^{1,2}$, Weibin Liu$^{1,2}$\thanks{Corresponding author: Weibin Liu, Email address: \texttt{wbliu@bjtu.edu.cn}}, Weiwei Xing$^3$\\
  $^1$ School of Computer Science and Technology, Beijing Jiaotong University \\
  $^2$ Visual Intelligence +X International Cooperation Joint Laboratory of MOE \\
  $^3$ School of Software Engineering, Beijing Jiaotong University\\
}

\begin{document}

\maketitle

\begin{abstract}
\label{sec:abs}

Vision-language models (VLMs) like CLIP have shown great success in image tasks, however, applying CLIP to open-vocabulary video recognition poses challenges. Due to the limited text information, directly fine-tuning CLIP on video recognition datasets may lead to overfitting and hinder its generalizability.
A prevalent method is to enrich text information by generating prompts using Large Language Models (LLMs). However, limited efforts have been made to explore how to generate higher-quality prompts that can contribute to classify unseen categories, and how to further enhance vision-text alignment based on these prompts. In this paper, we propose \textbf{P2CLIP} (\textit{Prompt-guided CLIP}), which aims to bridge this gap by exploring the impact of prompts across three aspects: category, video, and frame.At the category level, we argue that LLM-generated prompts under simple instructions may suffer from semantic overgeneralization, which can blur the category boundaries and hinder the performance of the model. To address this issue, we design Generic-Discriminative Prompt using LLMs to obtain discriminative category embeddings with rich semantic information. At the video and frame levels, we first use VLM and LLM to generate frame and video descriptions for enhancing vision-text alignment. Then we propose a Knowledge-Guided Temporal Aggregation mechanism, which leverages the expert knowledge derived from LLMs to guide the temporal modeling of videos and encourage a better construction of the joint visual-textual feature space. Extensive experiments across various benchmarks, including zero-shot and few-shot settings, demonstrate the effectiveness of our method.


%
\end{abstract}

\input{01-Introduction}
\input{02-RelatedWork}
\input{03-Method}

\input{04-Experiments}

\input{05-Conclusion}

{
\small
\bibliographystyle{plainnat}
\bibliography{egbib}
}

\clearpage
\input{06-Appendix}

\clearpage
\input{checklist}
\end{document}

%% file: 01-Introduction.tex
\section{Introduction}
\label{sec:intro}

Open-vocabulary learning requires the model to recognize categories beyond a predefined closed-set, which holds significant value in real-world applications. Recently, Vision-Language Models (VLMs) \cite{jia2021scaling, radford2021learning, yu2022coca, yuan2021florence} such as CLIP \cite{radford2021learning}, have shown great success due to its remarkable generalization capability, enabling image recognition under open-vocabulary settings. However, applying CLIP to open-vocabulary video recognition poses challenges. Due to the limited text information in the category names, (\eg, only 400 category names in Kinetics-400 \cite{carreira2017quo} and all are words or phrases), directly finetuning CLIP on the video recognition datasets may lead to overfitting and hinder its inherent generalizability.

To enrich the text information, a recently emerging method is utilizing Large Language Models (LLMs) to generate detailed and context-rich text descriptions. FROSTER \cite{huang2024froster} adopts GPT-3.5 to generate descriptions of category names
using the instruction: ``\textit{Please describe this action in the video \{\}.}''. Open-VCLIP++ \cite{wu2024building} uses BLIP-2 \cite{li2023blip} and Llama-2 \cite{touvron2023llama} to generate video caption for each video. Although using LLMs to generate detailed text descriptions has shown promise, limited efforts have been made to explore how to generate higher-quality prompts which can contribute to classify unseen categories, and how to further enhance vision-text alignment based on these prompts. In this paper, we propose \textbf{P2CLIP} (\textit{Prompt-guided CLIP}), which aims to bridge this gap by exploring the impact of natural language prompts across three aspects: category, video, and frame.


\begin{wrapfigure}{r}{0.49\textwidth}
\vspace{-0.2cm}
    \centering
    \begin{minipage}{0.49\textwidth}
        \includegraphics[width=1.0\textwidth]{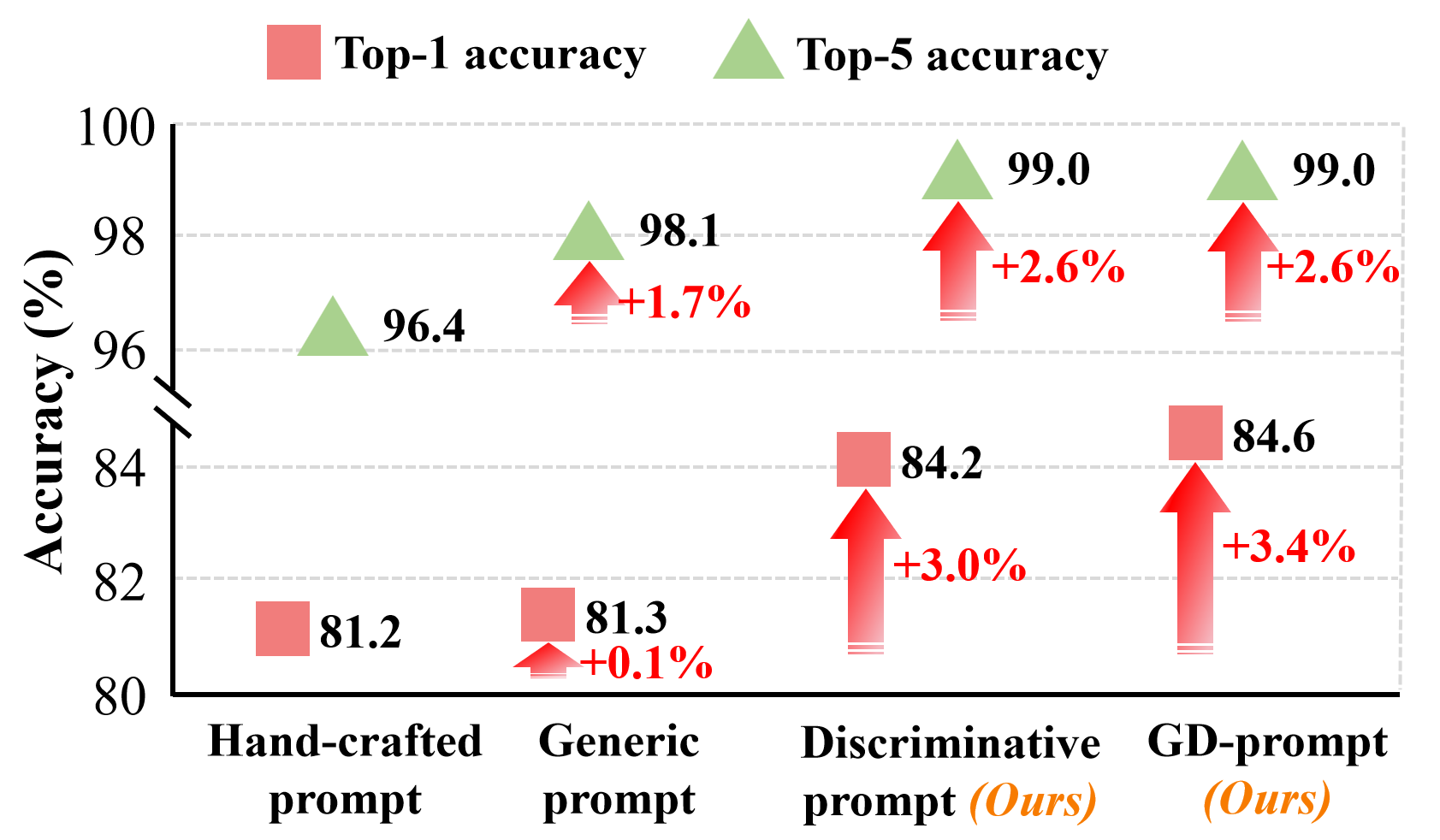}
        \caption{\textbf{
    Performance comparison} (Top-1 and Top-5 accuracy ($\%$)) under open-vocabulary setting where the model is tuned on Kinetics-400 \cite{carreira2017quo}, but evaluated on UCF-101 \cite{soomro2012ucf101}.}
    \label{fig1}
    \vspace{+0.3cm}
    \end{minipage}
    \vspace{+0.3cm}
    \begin{minipage}{0.49\textwidth}
        \includegraphics[width=1.0\textwidth]{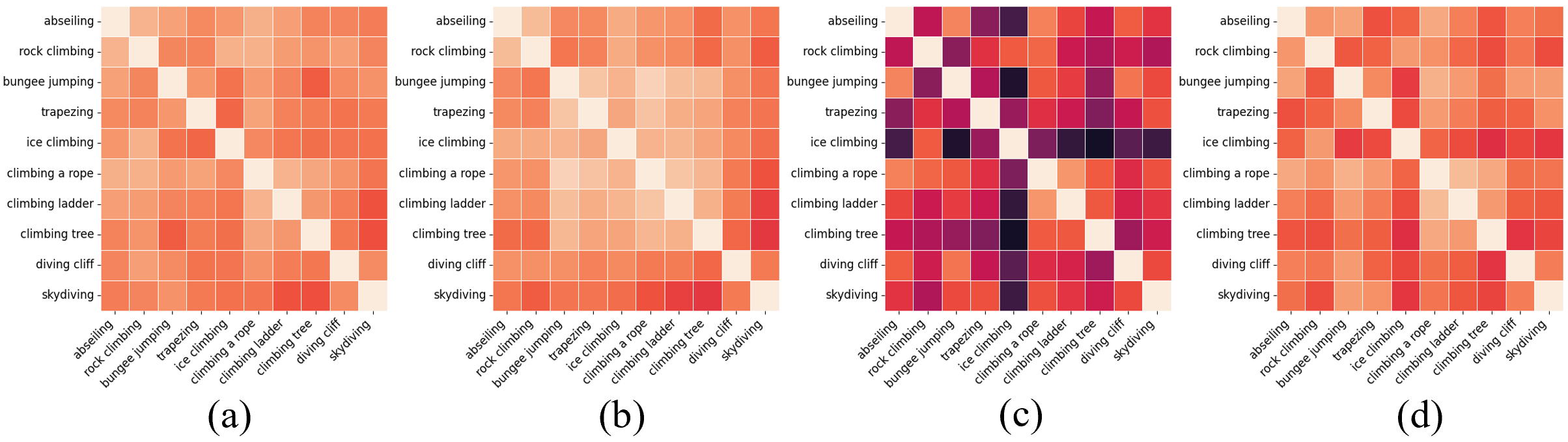}
        \caption{Inter-class correlation maps of 10 semantically similar categories in Kinetics-400. The darker the color, the lower the similarity. (a) Hand-crafted prompt. (b) Generic prompt. (c) Discriminative prompt. (d) GD-prompt.
        }
    \label{fig2}
    \vspace{-0.2cm}
    \end{minipage}
    \vspace{-0.4cm}
\end{wrapfigure}
At the category level, we argue that LLM-generated prompts under simple instructions (\ie, simply ask LLMs to describe a word) may suffer from semantic overgeneralization, which can blur the category boundaries. We conduct experiment to validate this hypothesis, as shown in Fig. \ref{fig1}. We adopt the \textit{Interpolated Weight Optimization} (IWR) and \textit{Stochastic Weight Averaging} (SWA) techniques in \cite{wu2024building} and the Hand-crafted prompt ``\textit{A video about \{\}.}'' to transform CLIP into a strong baseline.
Then, we employ the prompts in \cite{wu2023gpt4vis} as Generic prompt, which are generated by GPT-4 using simple instructions and contain a lot of detailed information. 
We observe that the model using Generic prompt improves Top-5 accuracy by 1.7$\%$ over baseline, but the Top-1 accuracy improved by only 0.1$\%$. The reason is that the detailed information in the Generic prompt expands the text space and mitigates overfitting, resulting in the improved prediction results across the entire category set. However, the semantic overgeneralization of the Generic prompt leads to more blurred semantic boundary between the semantically similar categories, resulting in very limited Top-1 accuracy improvement. To further validate, we use the textual encoder of CLIP to extract the text embeddings of 10 semantically similar categories with different prompts and calculate the correlation between them. The plot is shown in Fig. \ref{fig2}. Generic prompt significantly increases the inter-class similarities, blurring the semantic boundary between categories.

Therefore, we suppose that the LLM-generated prompts should contain rich detailed information, while increase the discriminability of the categories. Following this motivation, we employ LLMs to devise Discriminative prompt, which significantly decreases the inter-class similarities, as shown in Fig. \ref{fig2}
(c). Results in Fig. \ref{fig1} show that Discriminative prompt increases the Top-5 accuracy by 2.6$\%$, and Top-1 accuracy by 3.0$\%$. Finally, we combine the \textit{\textbf{G}eneric prompt} and \textit{\textbf{D}iscriminative prompt} to propose \textbf{\textit{GD-prompt}}, which is used in our method. As we can see in Fig. \ref{fig1}, GD-prompt increases Top-1 accuracy further by 0.4$\%$.

We further explore the impact of natural language prompt at video and frame levels.
Aligning the video feature with its corresponding video description has shown effectiveness for vision-text alignment in \cite{wu2024building}. In this paper, we improve the description generation method by using higher-performance VLM and LLM to generate fine-grained frame descriptions and video description for each input video. Furthermore, we note the limitations of video-description aligning process in \cite{wu2024building}: 
i) It only treats frame descriptions as intermediate results, overlooking the greater potential of the rich semantic information of them.
ii) Aligning the video feature obtained through meanpooling with its corresponding video description can lead to a suboptimal alignment of visual and textual knowledge.



\begin{figure}[!t]
  \centering
  \vspace{-0.2cm}
\includegraphics[width=1.0\textwidth]{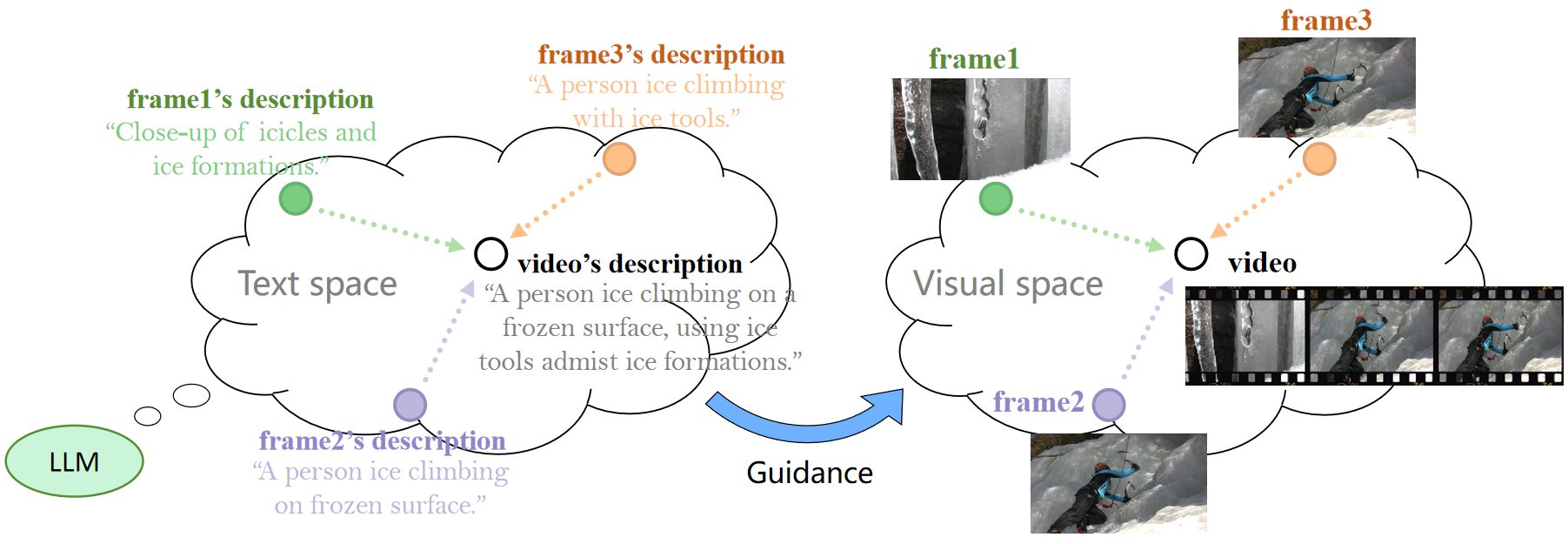}
  \caption{Visualization of the TGTA. We leverage the expert knowledge derived from LLMs to guide the temporal modeling of videos.
  }
  \label{fig3}
\vspace{-0.4cm}
\end{figure}

To address these limitations, we propose the \textit{\textbf{K}nowledge-\textbf{G}uided \textbf{T}emporal \textbf{A}ggregation mechanism} (\textbf{KGTA}) to further enhance vision-text alignment leveraging the knowledge derived from LLMs.
As illustrated in Fig. \ref{fig3}, the relationship between each frame description and the video description should be consistent with the relationship between the corresponding frame and the video.
Therefore, based on the expert knowledge of LLMs, we can leverage the reasoning process of LLMs in generating the video description as a guidance signal for the aggregation of frames.
Specifically, we use the similarity between the frame description and video description as the temporal saliency of the corresponding frame. By aligning the knowledge-guided video feature with the LLM-generated video description, we can encourage a better construction of the joint visual-textual feature space.

Our contributions can be summarized as follows:




\begin{itemize}
\item We propose \textit{Generic-Discriminative prompt} using LLMs to generate text descriptions of categories which contain rich detailed information while increase the discriminability of the categories. 
\item We use VLM and LLM to generate fine-grained frame and video descriptions. Building upon this, we propose \textit{Knowledge-Guided Temporal Aggregation mechanism} leveraging the expert knowledge derived from LLM to guide the temporal modeling of videoS, thereby further enhancing vision-text alignment.
\item We conduct extensive experiments on three popular video datasets, including UCF-101, HMDB-51, and Kinetics-600, to demonstrate the effectiveness of our method across various benchmarks including zero-shot and few-shot settings.
\end{itemize}

%% file: 02-RelatedWork.tex
\section{Related Work}
\label{sec:formatting}

\textbf{CLIP-based Video Recognition}
The rise of Vision-Language Models (VLMs) \cite{jia2021scaling, radford2021learning, yu2022coca, yuan2021florence}, such as CLIP\cite{radford2021learning}, have significantly advanced video recognition task due to its strong generalization capability. CLIP-based video recognition methods can be divided into two categories: full finetuning and efficient finetuning. Full finetuning methods \cite{wang2021actionclip, ni2022expanding, wu2023bidirectional, wu2023revisiting, tu2023implicit} update all parameters of the pre-trained CLIP model on the target dataset. Efficient finetuning methods \cite{ju2022prompting, pan2022st, yang2023aim, lin2022frozen, liu2023revisiting, wasim2023vita, wang2024m2, park2023dual, qing2023disentangling} focus on tuning only a small number of parameters or adding lightweight modules to the pre-trained CLIP, aiming to reduce computational costs and prevent overfitting, while still achieving competitive results. Although these methods have achieved notable progress, they often struggle in open-vocabulary tasks, where the model needs to recognize unseen categories during training.

Recently, researchers have begun to focus on the CLIP-based open-vocabulary video recognition. FROSTER \cite{huang2024froster}  uses frozen CLIP as the teacher model and designs a residual feature distillation module to supervise the feature learning process of the student model. Open-VCLIP \cite{weng2023open} treats the task as a continual learning problem without historical data and extends CLIP into an excellent zero-shot video classifier through the IWR and SWA techniques. As its follow-up work, Open-VCLIP++ \cite{wu2024building} utilizes BLIP-2 \cite{li2023blip} and Llama-2 \cite{touvron2023llama} to generate video descriptions, enhancing the vision-text alignment. AP-CLIP \cite{jia2024generating} generates action-conditioned prompts using LLMs and aligns videos with textual knowledge within the prompts. In this paper, we propose to generate context-rich and discriminative prompts using LLMs and leverage the expert knowledge embedded in LLMs to guide the aggregation of frames.

\noindent \textbf{Prompt Learning} is a parameter-efficient yet effective framework to finetune CLIP into downstream tasks. Prompt learning methods can be divided into soft prompt and hard prompt. Most soft prompt learning methods \cite{zhou2022learning, zhou2022conditional, goswami2024copl, ju2022prompting, yao2023visual, zhu2023prompt, bulat2023lasp} learn text prompt vectors. Other methods show the possibility of learning prompts in the visual space \cite{jia2022visual, bahng2022exploring} or in both visual and text space \cite{khattak2023maple, zang2022unified, wasim2023vita}. Early hard prompt learning methods \cite{wang2021actionclip, wu2023bidirectional} use Hand-crafted prompts, such as ``\textit{A video about \{\}.}''. Recently, some studies have explored to generate prompts using LLMs \cite{maniparambil2023enhancing, menon2022visual, pratt2023does, yang2023language, su2022language, yang2022empirical, pratt2023does}, which effectively align the visual feature with the rich textual knowledge embedded in LLMs. Inspired by the success of LLM-generated prompts in image tasks, there are also some works \cite{huang2024froster, weng2023open, wu2024building, wu2023gpt4vis, jia2024generating} in video recognition task that use LLMs to generate prompts. Most previous works directly use LLM-generated prompts under simple instructions, yet pay little attention to
what kind of prompts can truly aid in classifying unseen categories,
or how to further enhance vision-text alignment based on them. In this paper, we approach these problems 
by deeply exploring their impact across different tasks and how they can be further utilized to improve vision-text alignment.

%% file: 03-Method.tex
\section{Method}

\begin{figure*}[t]
  \centering
  \vspace{-0.3cm}
  \includegraphics[width=1.0\textwidth]{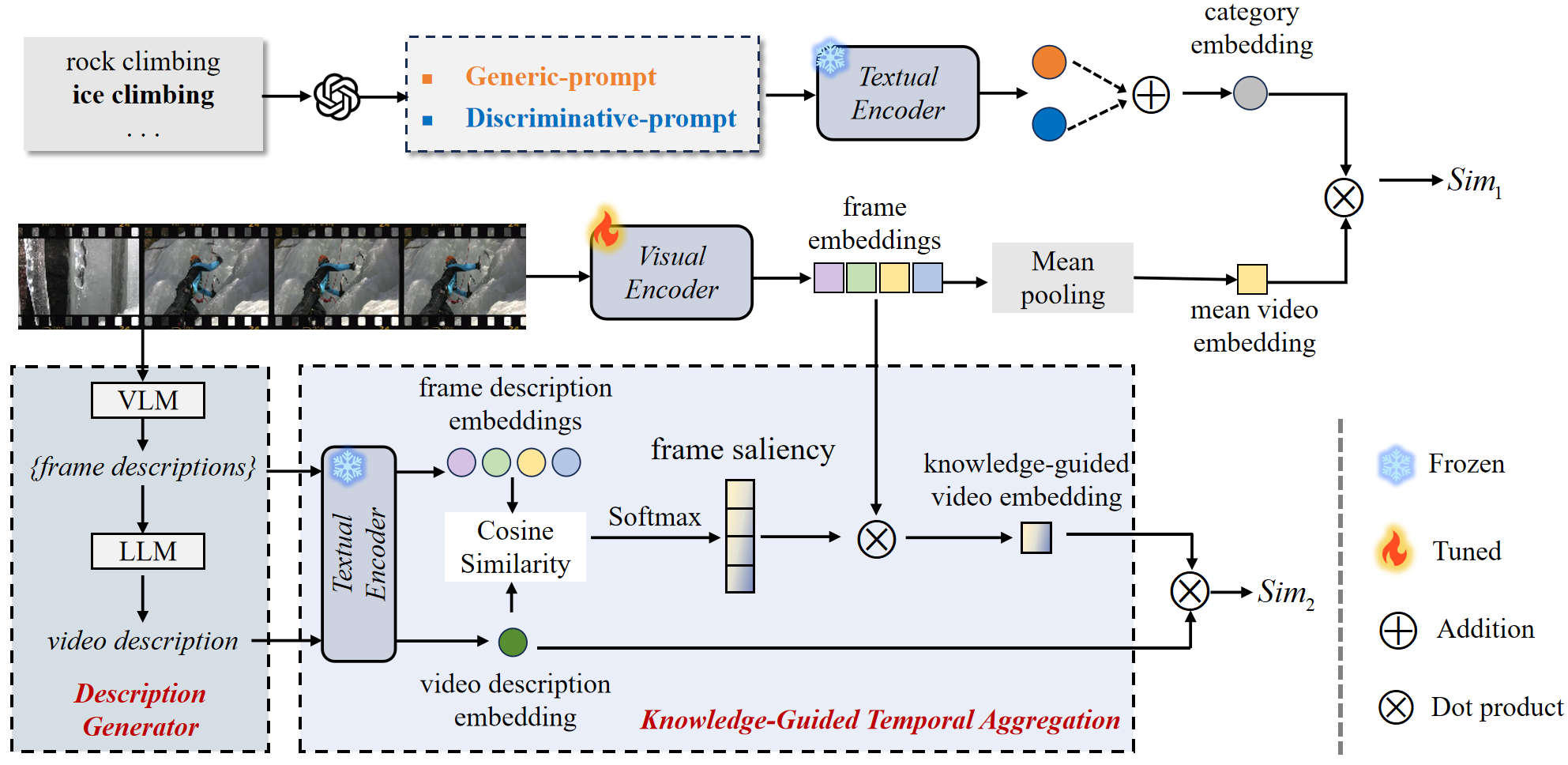}
  \caption{\textbf{Overview of P2CLIP for open-vocabulary video recognition.} We use the typical CLIP-based video recognition pipeline. We employ GPT4 to generate Generic prompt and Discriminative prompt for each category name and encode them to obtain GD-prompt embeddings. For each input video, we employ CogVLM2 and Llama-3.1 to generate frame and video descriptions, respectively, then we encode them into textual embeddings.
  We calculate the similarities between frame description embeddings and video description embedding as the temporal saliency of the corresponding frame.
  }
  \label{fig4}
  \vspace{-0.4cm}
\end{figure*}

An overview of our method for open-vocabulary video recognition is shown in Fig. \ref{fig4}. We elaborate on each component in this section. We begin with the typical CLIP-based video recognition pipeline in Sec. \ref{sec:3.1}. We then describe the Generic-Discriminative prompt (GD-prompt) generation in Sec. \ref{sec:3.2},  the frame and video description generator in Sec. \ref{sec:3.3}. Next, we propose the Knowledge-Guided Temporal Aggregation mechanism (KGTA) in Sec. \ref{sec:3.4} and the total loss function in Sec. \ref{sec:3.5}.

\subsection{CLIP-based Video Recognition}
\label{sec:3.1}

Given a video clip $V \in {\mathbb{R}^{T \times H \times W \times 3}}$ of spatial size $H \times W$ with $T$ sampled frames and a category name $C$, where $V$ and $C$ are sampled from a set of videos and a collection of category names respectively. We feed the video $V$ into the visual encoder $f\left( { \cdot |{\theta _v}} \right)$ and the category name $C$ into the textual encoder $f\left( { \cdot |{\theta _t}} \right)$ to obtain the visual embedding ${\mathbf{v}}$ and the category embedding ${\mathbf{c}}$, respectively.
\begin{equation}
    {\mathbf{v}} = f\left( {V|{\theta _v}} \right){\text{,       }}{\mathbf{c}} = f\left( {C|{\theta _t}} \right)
\end{equation}\label{eq1}
Then, we can use the cosine similarity to calculate the similarity score of ${\mathbf{v}}$ and ${\mathbf{c}}$ as follows:
\begin{equation}
    sim\left( {{\mathbf{v}},{\mathbf{c}}} \right) = \frac{{\left\langle {{\mathbf{v}},{\mathbf{c}}} \right\rangle }}{{\left\| {\mathbf{v}} \right\| \cdot \left\| {\mathbf{c}} \right\|}}
\end{equation}\label{eq1}

During training, our objective is maximizing the similarity score of ${\mathbf{v}}$ and ${\mathbf{c}}$ if $V$ and $C$ are matched, and otherwise minimizing it. The parameters ${\theta _v}$ and ${\theta _t}$ are initialized with weights from the pre-trained CLIP. During inference, we compute the similarity score between the give video and each category name, and choose the category with the highest similarity score as the top-1 prediction. 

\subsection{Generic-Discriminative Prompt Generation}
\label{sec:3.2}

Due to the limited text information, directly finetuning CLIP on the video recognition datasets may lead to overfitting and catastrophic forgetting. To expand the text space for improving results, some works use the Large Language Models (LLMs) to generate detailed text descriptions of category names, which has been demonstrated effective. We start by instructing LLMs to generate descriptive sentences as Generic prompt, which can describe the generic attributes of action categories. During actual operation, we employ the prompts of category names generated by GPT4 in \cite{wu2023gpt4vis} as Generic prompt. \textit{See appendix \ref{appendix_a} for more generation details and examples.}

\begin{wrapfigure}{r}{0.55\textwidth}
 \centering
  \vspace{-0.3cm}
\includegraphics[width=0.55\textwidth]{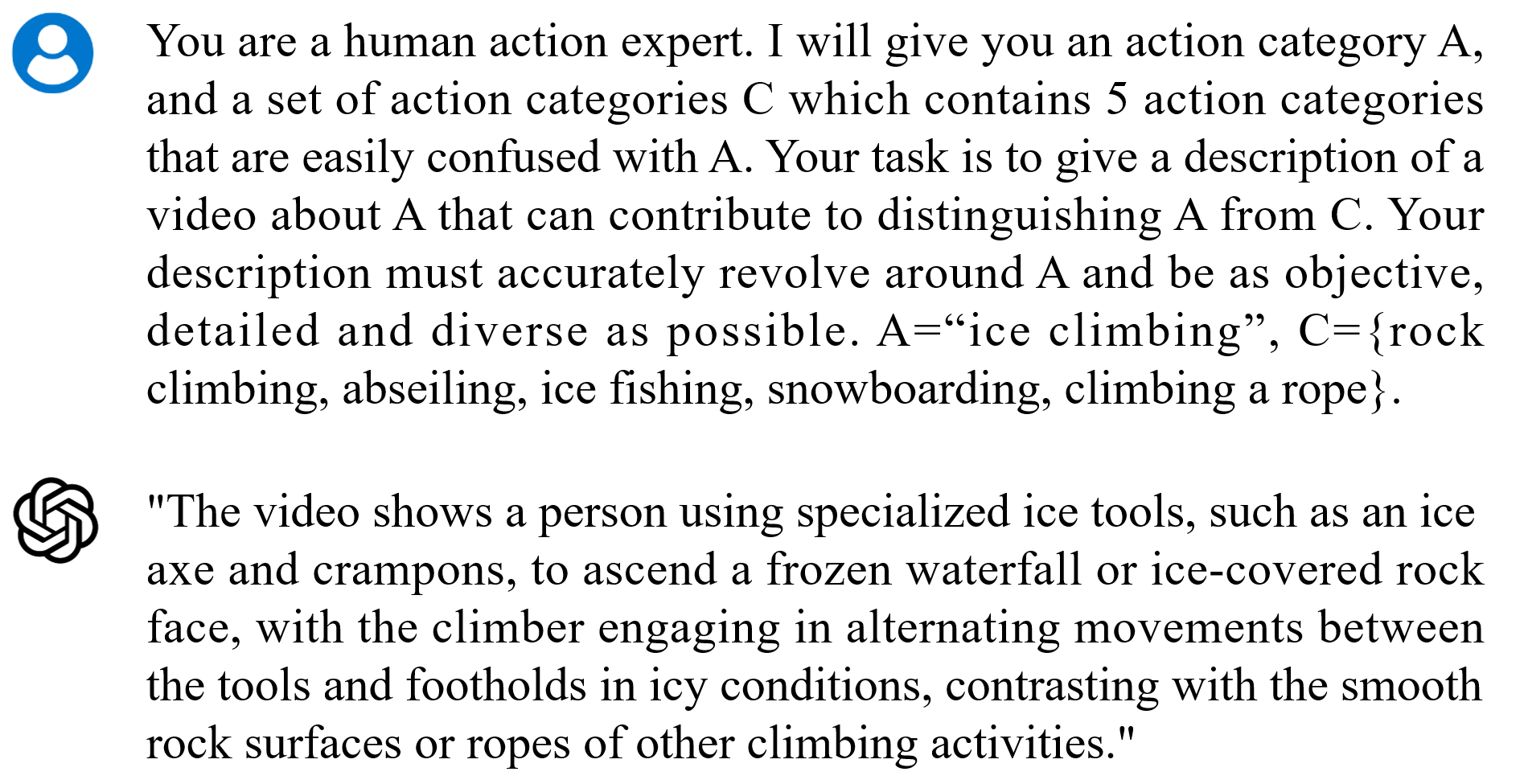}
  \caption{Instruction to generate Discriminative prompt and GPT4's response. 
  }
  \label{fig5}
\vspace{-0.4cm}
\end{wrapfigure}
For category name ``\textit{ice climbing}", its Generic prompt has 10 sentences with rich semantic information.
However, as discussed in Sec. \ref{sec:intro}, only using Generic prompt leads to a lack of discriminability between semantically similar categories, which is unsuitable for distinguishing a category from its semantic neighbor. Inspired by this, we propose to generate Discriminative prompt to obtain more discriminative category embeddings. First, we encode all category names into embeddings using CLIP's textual encoder. Then, we use cosine similarity to obtain its top-5 semantically similar categories, \ie, \textit{\{rock climbing, abseiling, ice fishing, snowboarding, climbing a rope\}}. Next, we generate its Discriminative prompt which can help to distinguish it from its semantic neighbor using GPT4. We present the instruction alongside GPT4's response in Fig. \ref{fig5}. \textit{See appendix \ref{appendix_b} for more examples.}

 Then, we concatenate it with the Hand-crafted prompt ``\textit{A video about {ice climbing.}}" and encode it using CLIP's textual encoder to get Discriminative prompt embedding ${{\mathbf{c}}^d}$ while the Generic prompt is also encoded and meanpooled to get ${{\mathbf{c}}^g}$. Then we can obtain the GD-prompt embedding as follows:
 \begin{equation}\label{equation3}
     {{\mathbf{c}}^{gd}} = \left( {{{\mathbf{c}}^g} + {{\mathbf{c}}^d}} \right)/2
 \end{equation}

GD-prompt contains detailed text information and increases discriminability between semantically similar categories. During training, we employ the cross-entropy loss to maximize the similarity score of ${{\mathbf{c}}^{gd}}$ and the meanpooled video embedding ${\mathbf{\bar v}}$ if they are matched, and otherwise minimize it. The loss function is as follows:
\begin{equation}\label{equation4}
    {\mathcal{L}_1} =  - \frac{1}{B}\sum\limits_{i = 1}^B {{\text{log}}\frac{{{\text{exp}}\left( {sim\left( {{{\mathbf{\bar{v}}}_i},{\mathbf{c}}_{true}^{gd}} \right)/\tau } \right)}}{{\sum\limits_{j = 1}^M {{\text{exp}}\left( {sim\left( {{{\mathbf{\bar{v}}}_i},{\mathbf{c}}_j^{gd}} \right)/\tau } \right)} }}} 
\end{equation}
where $B$ is batchsize, $M$ is the number of categories, ${{\mathbf{c}}_{true}^{gd}}$ is the GD-prompt embedding of the ground-truth category of ${{{\mathbf{\bar{v}}}_i}}$, $\tau$ is temperature hyperparameter.

\subsection{Frame and Video Description Generator}
\label{sec:3.3}

To align the visual content with more detailed text descriptions to combat overfitting, we generate fine-grained frame descriptions and video description for each input video using VLM and LLM which have excellent expert knowledge.

For video clip $V$ with $T$ sampled frames $\left\{ {{v_1},{v_2},...,{v_T}} \right\}$, we employ a powerful VLM, \ie, CogVLM2 \cite{wang2023cogvlm}, as the frame description generator to get $T$ frame descriptions $\left\{ {t_1^f,t_2^f,...,t_T^f} \right\}$ with the instruction:

\begin{quote}
    \textit{``In one sentence, describe the detail and movement of objects, and the action and pose of persons in the image. Prioritize accuracy."} 
\end{quote}


Then, we employ Llama-3.1 \cite{dubey2024llama} as the video description generator to aggregate the frame descriptions into a complete sentence that can appropriately describe the video $V$. The instruction is as follows:

\begin{quote}
\textit{``These are descriptions of the frames in temporal order of a video: \{\}. Please generate the description of the whole video according to the frame descriptions in one sentence. Start with ``The video shows".}
\end{quote}

These video-specific descriptions can significantly expand text space with their detailed text information and facilitate the vision-text alignment by aligned with visual concepts. \textit{See appendix \ref{appendix_c} for examples.}

\subsection{Knowledge-Guided Temporal Aggregation}
\label{sec:3.4}

Given a video clip $V$ with $T$ sampled frames $\left\{ {{v_1},{v_2},...,{v_T}} \right\}$, we get $T$ frame descriptions $\left\{ {t_1^f,t_2^f,...,t_T^f} \right\}$ and a video description ${t^v}$. Then we encode them using CLIP's textual encoder to get the frame description embeddings ${{\mathbf{c}}^f} \in {\mathbb{R}^{T \times d}}$ and video description embedding ${{\mathbf{c}}^v} \in {\mathbb{R}^{d}}$, where $d$ is the dimension of embedding. We calculate the similarity between each frame description embedding and the video description embedding to measure the relevance. Based on the expert knowledge of LLMs, we can regard the relevance as a guidance signal for the significance of each frame in the whole video. Then we perform a softmax operation to normalize the similarities and obtain the final temporal saliency for the $i$-th frame:
\begin{equation}
    {S_i} = \frac{{{\text{exp}}\left( {sim\left( {{\mathbf{c}}_i^f,{{\mathbf{c}}^v}} \right)/\tau } \right)}}{{\sum\limits_{j = 1}^T {{\text{exp}}\left( {sim\left( {{\mathbf{c}}_j^f,{{\mathbf{c}}^v}} \right)/\tau } \right)} }}
\end{equation}
where $\tau$ is the temperature hyperparameter. Then, we utilize the temporal saliency to aggregate the frame embeddings ${\mathbf{v}} \in {\mathbb{R}^{T \times d}}$ output from CLIP's visual encoder to get the knowledge-guided video embedding $\mathbf{\hat v}$ as follows:
\begin{equation}
    {\mathbf{\hat v}} \in {\mathbb{R}^d} = \sum\limits_{i = 1}^T {{\mathbf{v}}_i{S_i}} 
\end{equation}

During training, we employ the symmetric cross-entropy loss to maximize the similarity score of the knowledge-guided video embedding $\mathbf{\hat v}$ and the video description embedding ${{\mathbf{c}}^v}$ if they are matched, and otherwise minimize it:
\begin{equation}\label{equation7}
\begin{gathered}
  \mathcal{L}_2^{v2c} =  - \frac{1}{B}\sum\limits_{i = 1}^B {{\text{log}}\frac{{{\text{exp}}\left( {sim\left( {{{{\mathbf{\hat v}}}_i},{\mathbf{c}}_i^v} \right)/\tau } \right)}}{{\sum\limits_{j = 1}^B {{\text{exp}}\left( {sim\left( {{{{\mathbf{\hat v}}}_i},{\mathbf{c}}_j^v} \right)/\tau } \right)} }}}  \\ 
  \mathcal{L}_2^{c2v} =  - \frac{1}{B}\sum\limits_{i = 1}^B {{\text{log}}\frac{{{\text{exp}}\left( {sim\left( {{\mathbf{c}}_i^v,{{{\mathbf{\hat v}}}_i}} \right)/\tau } \right)}}{{\sum\limits_{j = 1}^B {{\text{exp}}\left( {sim\left( {{\mathbf{c}}_i^v,{{{\mathbf{\hat v}}}_j}} \right)/\tau } \right)} }}}  \\ 
  {\mathcal{L}_2} = \frac{1}{2}\left( {\mathcal{L}_2^{v2c} + \mathcal{L}_2^{c2v}} \right) \\ 
\end{gathered} 
\end{equation}\label{eq7}
where $\tau$ is the temperature hyperparameter, $B$ is batchsize. \textit{See appendix \ref{appendix_c} for visualization of temporal saliencies.}

\subsection{Total Loss Function}
\label{sec:3.5}
After obtaining the loss $\mathcal{L}_1$ in Eq. \ref{equation4} and the loss $\mathcal{L}_2$ in Eq. \ref{equation7}, the total loss function during training is the combination of $\mathcal{L}_1$ and $\mathcal{L}_2$ as follows:
\begin{equation}\label{equation8}
    \mathcal{L} = {\mathcal{L}_1} + \lambda {\mathcal{L}_2}
\end{equation}\label{eq8}
where $\lambda$ is a weight coefficient. The loss $\mathcal{L}_1$ aligns the meanpooled video embedding with the augmented category embedding to learn the distribution of category labels. The loss $\mathcal{L}_2$ aligns the knowledge-guided video embedding with the LLM-generated video description for a better alignment of visual and textual knowledge.

%% file: 04-Experiments.tex
\section{Experiments}


\subsection{Experimental Settings}
\label{sec:4.1}

\noindent\textbf{Datasets.} We conduct experiments use four popular video datasets, \ie, Kinetics-400\cite{carreira2017quo}, UCF-101\cite{soomro2012ucf101}, HMDB-51\cite{kuehne2011hmdb} and Kinetics-600\cite{carreira2018short}. \textit{See appendix \ref{appendix_d} for details of these datasets.}

\noindent\textbf{Implementation Details.} 

(i) \textit{Zero-shot setting:} We adopt the IWR and SWA techniques in \cite{wu2024building} with Hand-crafted prompt ``\textit{A video about \{\}}.'' to transfer CLIP into a strong baseline. Then, we equip the baseline with GD-prompt, description generator and KGTA to build the open-vocabulary video model. We use the Kinetics-400 dataset for training and conduct testing on UCF-101, HMDB-51, along with a subset of the Kinetics-600 dataset. During training, we use all three proposed components, while during testing, we only use the GD-prompt of the target dataset. We evaluate the model following the evaluation protocols widely used in previous works. \textit{See appendix \ref{appendix_d} for more details on evaluation protocols.}

(ii) \textit{Few-shot setting:} We begin with the model variant with frozen textual encoder in \cite{rasheed2023fine} as the baseline. Note that we equip the baseline with only GD-prompt, which we find more effective for few-shot tasks and we will discuss this in Sec. \ref{sec4.3}. We directly perform finetuning on the target dataset, with no pre-training on Kinetics-400.

We extract 8 frames for each video and use ViT-B/16 based CLIP model for all experiments. During testing, we use three clips with one crop (``3$\times$1 views''). The weight coefficient $\lambda$ of the total loss function is 2.0. \textit{See appendix \ref{appendix_e} for more training details.}



\subsection{Comparison with the State-of-the-art}
\label{sec:4.2}

\subsubsection{Zero-shot Setting}
\label{sec:4.2.1}

In zero-shot setting, we demonstrate the capability of our method to recognize unseen categories during training, showcasing its generalizability in open-set scenarios.
We compare our method with other CLIP-based methods on UCF-101 and HMDB-51 under \textit{Evaluation Protocol 1} and \textit{Evaluation Protocol 2} in Table \ref{tab:1}, under \textit{Evaluation Protocol 3} in Table \ref{tab:2}. Our method consistently outperforms other methods across all evaluation protocols, for example, our method surpasses Open-VCLIP++ by +1.1$\%$ on UCF-101, and +0.8$\%$ on HMDB-51 under \textit{Evaluation Protocol 1}. When considering the Kinetics-600 dataset, as illustrated in Table \ref{tab:3}, our method also outperforms most of the other CLIP-based methods by a clear margin. P2CLIP consistently achieves the best performance across all datasets and all evaluation protocols, highlighting its open-vocabulary recognizing capability.

\subsubsection{Few-shot Setting}
\label{sec:4.2.2}

In few-shot setting, we extend our method to supervised scenarios where the training data is extremely limited.
We compare our method to various CLIP-based methods in Table \ref{tab:4}. As we can see, our method achieves impressive performance and consistently outperforms all other methods on both UCF-101 and HMDB-51 datasets across all shot divisions (\ie, $K=2,4,8,16$). On HMDB-51 dataset, by simply adopting GD-prompt, our method gets +3.8$\%$, +3.2$\%$, +4.0$\%$, +1.9$\%$ in 2, 4, 8, 16 shots, respectively.
Additionally, our method tends to show greater gains when the training data is more limited, highlighting that GD-prompt can compensate for the lack of visual information by providing rich text knowledge, while also enhancing the model's understanding of action categories by linking it textual information to the knowledge in the pre-trained model.

\vspace{0.1cm}
\begin{table}[!t]
  \caption{Zero-shot classification performance for CLIP-based methods on UCF-101 and HMDB-51 under EP1 and EP2. The best results are \textbf{bolded}, and the second-best results are \underline{underlined}.
  }
  \vspace{-0.1cm}
  \label{tab:1}
  \centering
  \begin{small}
  \begin{tabular}{lccccc}
    \toprule[1pt]
    \rule{0pt}{10pt}
    \multirow{2}{*}{\makebox[0.13\textwidth][l]{Method}} & \multirow{2}{*}{\makebox[0.13\textwidth][c]{Encoder}} & \multicolumn{2}{c}{UCF-101} & \multicolumn{2}{c}{HMDB-51} \\ \cline{3-6} 
    
    \rule{0pt}{10pt}
        & & \makebox[0.09\textwidth][c]{EP1} & \makebox[0.09\textwidth][c]{EP2} & \makebox[0.09\textwidth][c]{EP1} & \makebox[0.09\textwidth][c]{EP2} \\ 
    \midrule
    ActionCLIP \cite{wang2021actionclip}& ViT-B/16 & - & 69.5 & - & 50.5 \\
    Text4Vis \cite{wu2023revisiting}& ViT-L/14 &  85.8$ \pm $3.3 & 79.6 &  58.1$ \pm $5.7 & \\
    BIKE \cite{wu2023bidirectional} & ViT-L/14 &  86.6$ \pm $3.4 & 80.8 &  61.4$ \pm $3.6 & 52.8\\
    Open-VCLIP \cite{weng2023open} & ViT-B/16 &  89.9$ \pm $1.7 & 83.5 &  64.5$ \pm $4.5 & 53.2\\
    Open-VCLIP++ \cite{wu2024building} & ViT-B/16 &  \underline{90.0$\pm$1.7} & 84.1 &  \underline{65.8$\pm$4.3} & \underline{54.7}\\
    FROSTER \cite{huang2024froster} & ViT-B/16 & - & \underline{85.0} & - & 54.5\\
    \midrule
    \rowcolor{gray!20} 
    P2CLIP (ours)  & ViT-B/16 &  \textbf{91.1$\pm$1.5} & \textbf{85.1} & \textbf{66.6$\pm$5.0} & \textbf{55.2} \\
    \bottomrule[1pt]
  \end{tabular}
  \end{small}
\vspace{-0.25cm}
\end{table}


\begin{figure}[!t]
    \centering
    \begin{minipage}[t]{0.5\textwidth}
        \centering
        \captionof{table}{Zero-shot classification performance for CLIP-based methods on UCF-101 and HMDB-51 under EP3.}
        \vspace{-0.1cm}
        \label{tab:2}
        \begin{small}
        \begin{tabular}{lcc}
    \toprule[1pt]
    \rule{0pt}{10pt}
    \makebox[0.15\textwidth][l]{Method}  & \makebox[0.1\textwidth][c]{UCF-101} & \makebox[0.1\textwidth][c]{HMDB-51} \\
    \midrule
    ActionCLIP \cite{wang2021actionclip}&  58.3$ \pm $3.4 & 40.8$ \pm $5.4 \\
    XCLIP \cite{ni2022expanding}& 72.0$ \pm $2.3 & 44.6$ \pm $5.2 \\
    ViFi-CLIP \cite{rasheed2023fine} & 76.8$ \pm $0.7 & 51.3$ \pm $0.6 \\
    M$^2$-CLIP\cite{wang2024m2} & 78.7$ \pm $1.2 & 47.1$ \pm $0.4 \\
    AP-CLIP \cite{jia2024generating} &   82.4$ \pm $0.5 & 55.4 $\pm$ 0.8 \\
    Open-VCLIP \cite{weng2023open} &  83.4$ \pm $1.2 & 53.9$ \pm $1.2 \\
    Open-VCLIP++ \cite{wu2024building} &   83.9$ \pm0.6$ & \underline{55.6$\pm$1.4} \\
    FROSTER \cite{huang2024froster} &   \underline{84.8$\pm$1.1} & 54.8$ \pm $1.3 \\
    \midrule
    \rowcolor{gray!20}
    P2CLIP (ours) &  \textbf{85.1$\pm$0.1} & \textbf{55.7$\pm$1.0} \\
    \bottomrule[1pt]
  \end{tabular}
  \end{small}
    \end{minipage}%
    \hspace{0.5cm} 
    \begin{minipage}[t]{0.35\textwidth}
        \centering
        \captionof{table}{Zero-shot classification performance for CLIP-based methods on Kinetics-600.}
        \vspace{-0.1cm}
        \label{tab:3}
        \begin{small}
        \begin{tabular}{lc}
    \toprule[1pt]
    \rule{0pt}{10pt}
    \makebox[0.15\textwidth][l]{Method}  & \makebox[0.1\textwidth][c]{K-600}  \\
    \midrule
    Vanilla CLIP\cite{radford2021learning} &59.8$ \pm $0.3 \\
    ActionCLIP \cite{wang2021actionclip}&  66.7$ \pm $1.1 \\
    XCLIP \cite{ni2022expanding}& 65.2$ \pm $0.4  \\
    A5\cite{ju2022prompting} &55.8$ \pm $0.7  \\
    ViFi-CLIP \cite{rasheed2023fine} & 71.2$ \pm $1.0 \\
    AP-CLIP \cite{jia2024generating} &   73.4$ \pm $0.8  \\
    Open-VCLIP \cite{weng2023open} &  73.0$ \pm $1.0\\
    Open-VCLIP++ \cite{wu2024building} &   \underline{73.4$\pm$0.8}\\
    
    \midrule
    \rowcolor{gray!20}
    P2CLIP (ours) &  \textbf{74.8$\pm$0.8}  \\
    \bottomrule[1pt]
  \end{tabular}
  \end{small}
    \end{minipage}
    \vspace{-0.25cm}
\end{figure}

\begin{table*}[!ht]
\centering
  \caption{Few-shot performance for CLIP-based methods on UCF-101 and HMDB-51. We employ the variant model of ViFi-CLIP with frozen textual encoder as a baseline and simply integrate GD-prompt into it. \textsuperscript{†} denotes the results of our reproduction. Gains in \textcolor{green_im}{green} are compared to baseline. 
  }
  \vspace{-0.1cm}
  \label{tab:4}
  \centering
  \begin{small}
  \setlength{\tabcolsep}{5pt}
  \begin{tabular}{lllll|llll}
    \toprule[1pt]
    \rule{0pt}{10pt}
    \multirow{2}{*}{\makebox[0.1\textwidth][l]{Method}}  & \multicolumn{4}{c}{UCF-101} & \multicolumn{4}{c}{HMDB-51} \\ \cline{2-9} 
    \rule{0pt}{10pt}
    &$K$=2 & $K$=4 & $K$=8 & $K$=16 & $K$=2 & $K$=4 & $K$=8 & $K$=16 \\
    \midrule
    ActionCLIP &  70.6 & 71.5 &73.0&91.4 & 47.5 & 57.9 & 57.3 &  59.1 \\
    XCLIP \cite{ni2022expanding}&  48.5 & 75.6 &83.7&91.4 & \underline{53.0} & 57.3 & 62.8 &  64.0\\
    A5 \cite{ju2022prompting}&  71.4 & 79.9 &85.7&89.9 & 39.7 & 50.7 & 56.0 &  62.4\\
    ViFi-CLIP\textsuperscript{†} \cite{rasheed2023fine}&  \underline{83.8} & \underline{86.8}& \underline{88.8}& \underline{92.1} & 52.9 & \underline{59.7} & \underline{65.4} & \underline{74.0}  \\
    \midrule
    \rowcolor{gray!20} 
    P2CLIP (ours) & \textbf{86.0\rise{2.2}}& \textbf{88.1\rise{1.3}}& \textbf{90.1\rise{1.3}}& \textbf{93.3\rise{1.2}}& \textbf{56.7\rise{3.8}}& \textbf{62.9\rise{3.2}}& \textbf{69.4\rise{4.0}} & \textbf{75.9\rise{1.9}}  \\
    \bottomrule[1pt]
  \end{tabular}
  \end{small}
  \vspace{-0.6cm}
\end{table*}

\subsection{Discussion and Ablation Studies}
\label{sec4.3}

\begin{figure}[t]
    \centering
    
    \begin{minipage}{0.45\textwidth}
        \centering
        \captionof{table}{Ablations for component.}
        \label{tab:5}
        \begin{small}
        \setlength{\tabcolsep}{2pt}
        \begin{tabular}{l|lll}
    \toprule[1pt]
    \rowcolor{gray!20} 
    \makebox[0.1\textwidth][l]{Method}  &\makebox[0.01\textwidth][l]{UCF}  & \makebox[0.01\textwidth][l]{HMDB} &\makebox[0.01\textwidth][l]{K600} \\
    \midrule
    baseline & 81.2 &  48.8 &72.3 \\
    + GD-prompt & 84.5\rise{3.3}& 53.9\rise{5.1}& 74.3\rise{2.0}\\
    + Video description & 85.0\rise{3.8}& 54.8\rise{6.0}& 74.6\rise{2.3}\\
    + KGTA & 85.1\rise{3.9}& 55.2\rise{6.4}& 74.8\rise{2.5}\\

    \bottomrule[1pt]
  \end{tabular}
\end{small}
\vspace{-0.1cm}
    \end{minipage}
    \hfill
    \begin{minipage}{0.5\textwidth}
        \centering
        \includegraphics[width=1.0\textwidth]{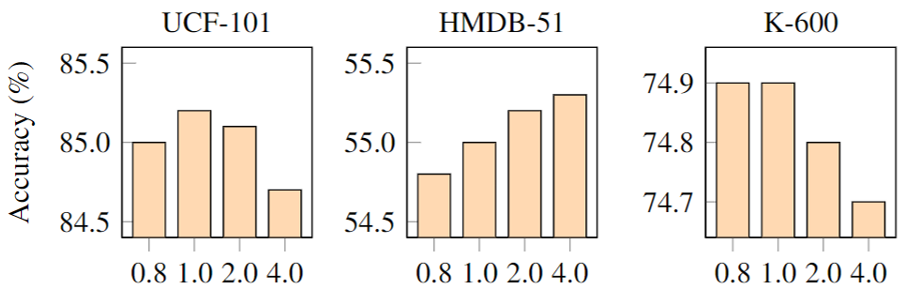}
  \caption{Analysis for $\lambda$ under zero-shot setting.}
  \label{fig6}
  \vspace{-0.1cm}
    \end{minipage}
    
    \vskip\baselineskip
    
    \begin{minipage}{0.45\textwidth}
        \centering
        \captionof{table}{Ablations for different text prompts.}
        \label{tab:6}
         \begin{small}
        \setlength{\tabcolsep}{2pt}
        \begin{tabular}{l|lll}
  \toprule[1pt]
    \rowcolor{gray!20} 
    \makebox[0.1\textwidth][l]{Prompt}  & \makebox[0.01\textwidth][l]{UCF} & \makebox[0.01\textwidth][l]{HMDB}&\makebox[0.01\textwidth][l]{K600}\\
    \midrule
    Hand-crafted prompt & 81.2 & 48.8 &72.3\\
    Generic prompt & 81.3\rise{0.1}  & 49.5\rise{0.8}&72.8\rise{0.5}\\
    Discriminative prompt & 84.2\rise{3.0} & 52.3\rise{3.5}&73.5\rise{1.2}\\
    GD-prompt & \textbf{84.5\rise{3.3}}  & \textbf{53.9\rise{5.1}} & \textbf{74.3\rise{2.0}}\\
    \bottomrule[1pt]
  \end{tabular}
  \end{small}
    \end{minipage}
    \hfill
    \begin{minipage}{0.5\textwidth}
        \centering
        \captionof{table}{Ablations for different description generators on UCF-101.}
        \label{tab:7}
        \begin{small}
        \setlength{\tabcolsep}{4pt}
  \begin{tabular}{l|c}
    \toprule[1pt]
    \rowcolor{gray!20} 
    \makebox[0.6\textwidth][l]{Method}  &\makebox[0.2\textwidth][c]{Top-1.($\%$)} \\
    \midrule
    Open-VCLIP\cite{weng2023open} & 83.5 \\
    +BLIP2/Llama2 (Open-VCLIP++\cite{wu2024building}) &84.1\rise{0.6}\\
    +CogVLM2/Llama3.1 &\textbf{84.6\rise{1.1}}\\
    \bottomrule[1pt]
  \end{tabular}
  \end{small}
    \end{minipage}

\vspace{-0.6cm}
\end{figure}

\noindent\textbf{Effectiveness of Components.} Table \ref{tab:5} demonstrates the effectiveness of our proposed components. 
With GD-prompt, we can obtain significant performance gains over baseline by +3.3$\%$, +5.1$\%$, +2.0$\%$ on UCF-101, HMDB-51, K-600, respectively, demonstrating that GD-prompt can effectively mitigate overfitting and aid classification with its discriminative text information.
Then, aligning the meanpooled video feature not only with its augmented category label but also with its corresponding video description can further gain +0.5$\%$, +0.9$\%$, +0.3$\%$ on UCF-101, HMDB-51, K-600, respectively.
Finally, we introduce the KGTA mechanism, 
by aligning the video description with the knowledge-guided video feature, rather than the meanpooled video feature, we encourage the model to construct a better joint visual-textual feature space. The KGTA mechanism brings +0.1$\%$, +0.4$\%$, +0.2$\%$ on UCF-101, HMDB-51, K-600, respectively, demonstrating its effectiveness.

\noindent\textbf{Ablations for Different Text Prompts.} Table \ref{tab:6} shows the effect of different text prompts. As we discussed in Sec. \ref{sec:intro}, Generic-prompt brings limited Top-1 accuracy improvements, especially on UCF-101 (only +0.1$\%$). 
When using Discriminative prompt, the model outperforms the baseline by a significant margin, demonstrating the effectiveness of the discriminative text descriptions, which are beneficial for classification.
When using GD-prompt, we combine the generic attributes of actions with the discriminative descriptions, and obtain +3.3$\%$, +5.1$\%$, +2.0$\%$ on UCF-101, HMDB-51, K-600, respectively, highlighting the effectiveness of the prompt strategy we propose. \textit{See appendix \ref{appendix_f} for more visualizations of inter-class correlation maps for different text prompts.}


\noindent\textbf{Ablations for Different Description Generators.} Table \ref{tab:7} shows the effect of different frame and video description generators on the UCF-101 dataset. We use Open-VCLIP\cite{weng2023open} as a baseline. Building upon it, Open-VCLIP++\cite{wu2024building} uses BILP-2 and Llama2 as the frame and video description generator, respectively. By aligning the visual feature with its corresponding video description, Open-VCLIP++ gains +0.6$\%$ on UCF-101. Instead, we use CogVLM2 and Llama3.1 as the frame and video description generator respectively, and gains +1.1$\%$, highlighting the effectiveness of using higher-performance VLM and LLM.

\noindent\textbf{Parameter Analysis of $\lambda$ under Zero-shot Setting.}  
Fig. \ref{fig6} shows the effect of different values of $\lambda$ (0.8, 1, 2, 4), which is the weight coefficient of the total loss function in Eq. \ref{equation8}. 
A larger $\lambda$ indicates that the model is encouraged to receive more supervision from video descriptions during training. 
On UCF-101 and K-600, the model achieves the best performance (85.2$\%$ and 74.9$\%$) when $\lambda$ is 1.0, as it getting larger, the performance begins to drop. 
On HMDB-51, as $\lambda$ getting larger, the performance of the model tends to increase (54.8$\%$, 55.0$\%$, 55.2$\%$, 55.3$\%$). We choose 2.0 as the final value of $\lambda$ in our method, which achieves 85.1$\%$, 55.2$\%$, 74.8$\%$ on UCF-101, HMDB-51, K-600, respectively.




\begin{wrapfigure}{r}{0.6\textwidth} 
\vspace{-0.4cm}
    \centering
    \begin{minipage}{0.3\textwidth}
        \centering
        \includegraphics[width=\linewidth]{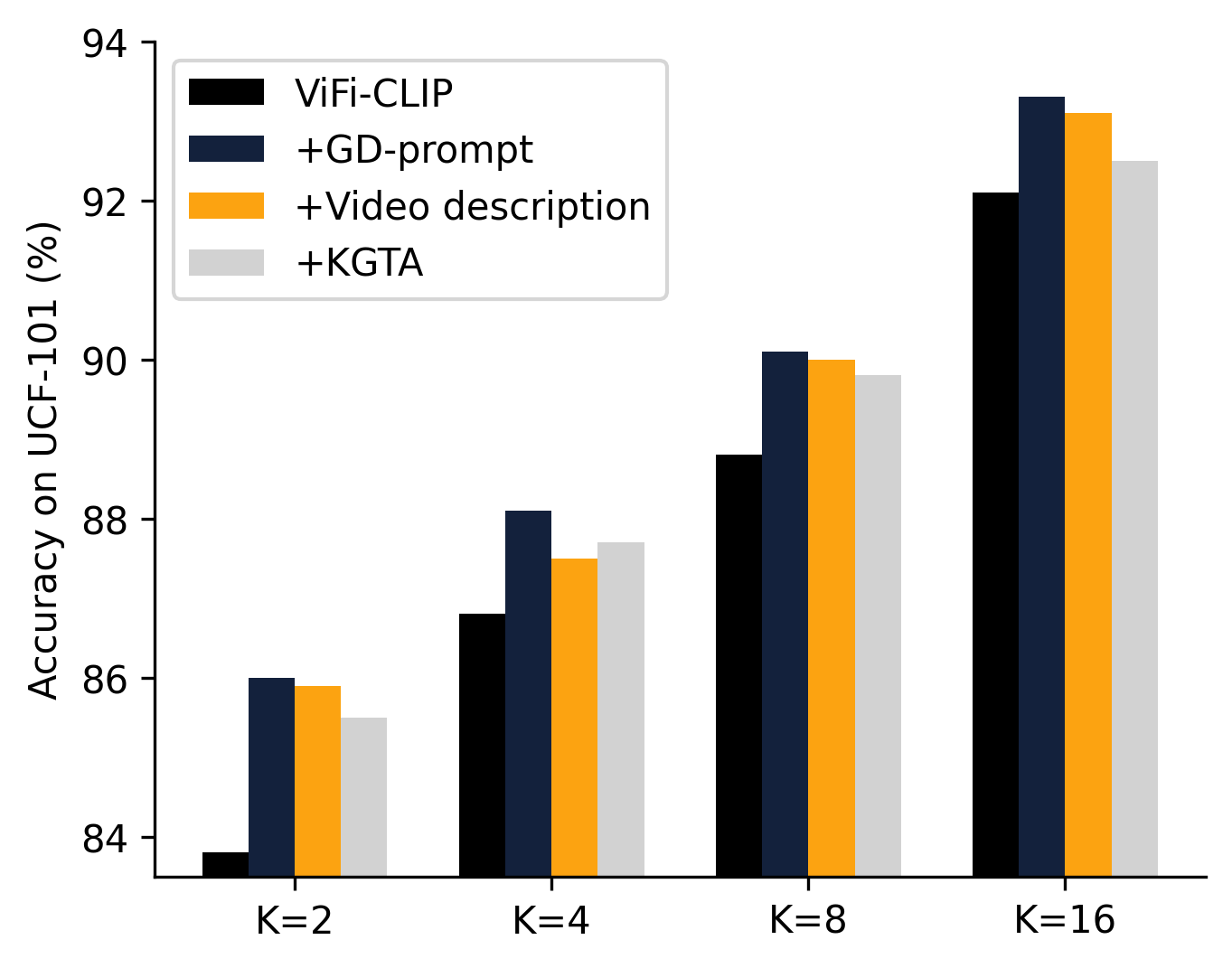}
        \vspace{-0.3cm}
    \end{minipage}%
    \begin{minipage}{0.3\textwidth}
        \centering
        \includegraphics[width=\linewidth]{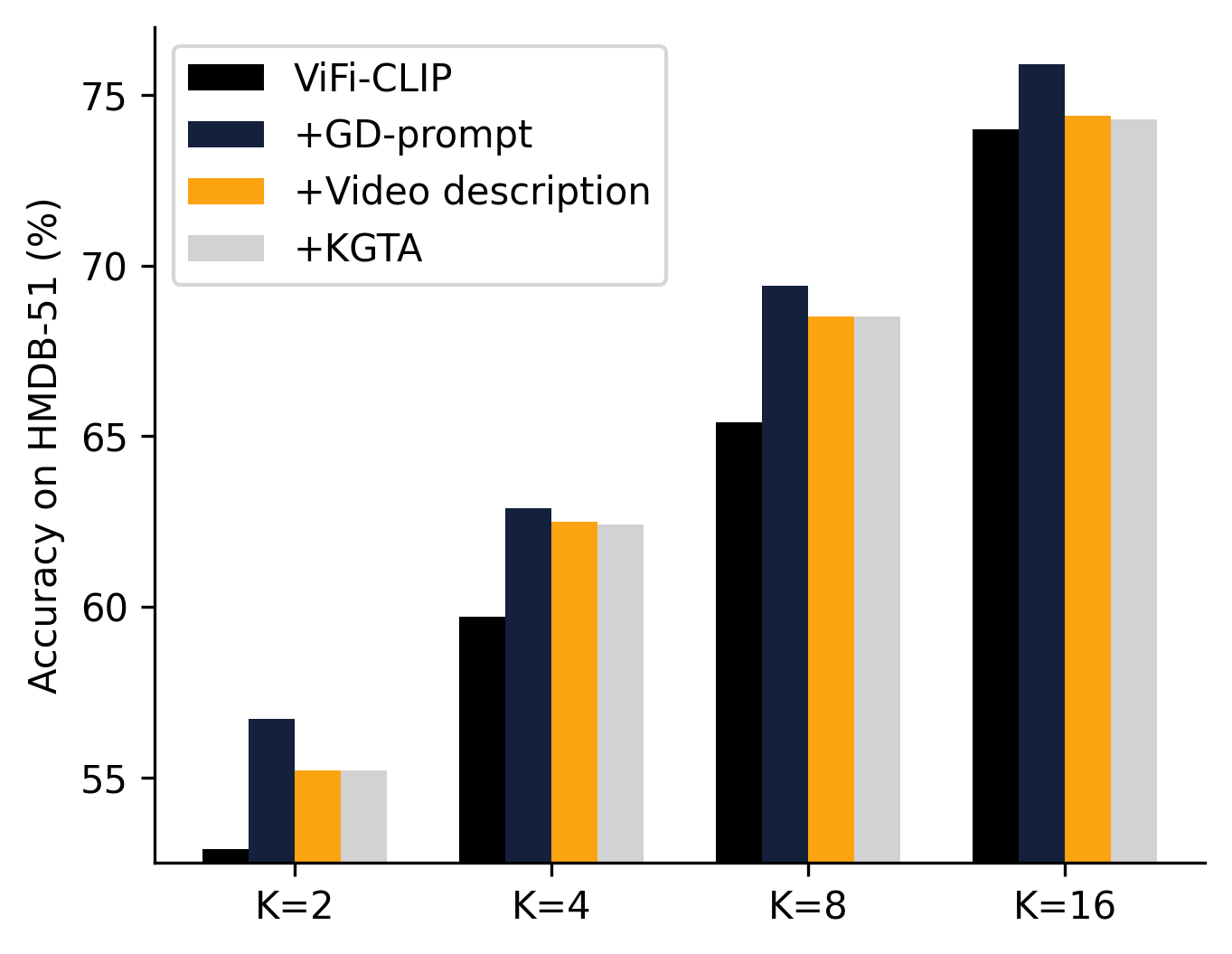}
        \vspace{-0.3cm}
    \end{minipage}
    \vspace{-0.4cm}
    \caption{Ablations for components on UCF-101 (\textbf{Left}) and HMDB-51 (\textbf{Right}) under few-shot setting.}
    \label{fig:7}
\vspace{-0.5cm}
\end{wrapfigure}
\noindent\textbf{Effect of Components under Few-shot Setting.} 
Fig. \ref{fig:7} shows the effect of proposed components on UCF-101 and HMBD-51 under few-shot setting.
Using GD-prompt brings consistent performance improvements across all shot divisions.
When we further introduce the video descriptions and KGTA, the performance of the model degrades by varying degrees. 
We hypothesize there are two reasons. 
First, in closed-set tasks, the core objective of the model is to learn the joint distribution of category labels, rather than the vision-text alignment information provided by video descriptions, which may distract the model's attention.
Second, in few-shot setting, the model's learning capacity is limited, preventing it from fully learning the vision-text alignment knowledge from such a small number of video-text pairs. As a result, it struggles to generalize well to other video data, especially when the LLM-generated descriptions do not fully match the data distribution or are complex, which could introduce noise and negatively affect the model's performance. 
Based on the above analysis, we only use GD-prompt in few-show setting, which can still achieve impressive performance.



%% file: 05-Conclusion.tex
\section{Conclusion}
\label{sec:conclu}

In this paper, we conduct extensive exploration of the LLM-generated natural language prompts across different tasks. First, we propose to generate Generic-Discriminative prompt which contains both context-rich and discriminative information that aids classification. Next, we generate high-quality frame and video descriptions for each video, enhancing vision-text alignment by aligning visual concepts with detailed descriptions. Based on these video-specific descriptions, we propose the Knowledge-Guided Temporal Aggregation mechanism, leveraging the expert knowledge embedded in LLMs to help temporal modeling and build a better joint visual-textual feature space. Extensive experiments demonstrate the effectiveness of our method.    

\noindent\textbf{Limitations.}
Although our method achieves good performance across all datasets, the impact of the parameter $\lambda$ varies when the test data has different domain gaps from the training data, which dedicates that different test data may have varying needs for supervision derived from video descriptions. Thus, a promising future research direction is to build a cross-domain open-vocabulary video classifier that can effectively adapt to the test data with different gaps from the training data. 

\section*{Acknowledgement}

This research is supported by the Beijing Natural Science Foundation (No. L231005).

%% file: 06-Appendix.tex
\appendix
\section*{Appendix}
\label{sec:Appendix}

This appendix offers extensive additional details and more qualitative analysis complementing the main paper. The content is organized as follows:
\begin{itemize}
\item Generic prompts in GPT4Vis (Appendix \ref{appendix_a})
\item More examples of Discriminative prompt (Appendix \ref{appendix_b})
\item Examples of frame and video descriptions with temporal saliency in KGTA (Appendix \ref{appendix_c})
\item Details of datasets and evaluation protocols in zero-shot setting (Appendix \ref{appendix_d})
\item More training details (Appendix \ref{appendix_e})
\item More visualizations of inter-class correlation maps for different text 
 prompts(Appendix \ref{appendix_f})
\end{itemize}

\section{Generic prompts in GPT4Vis}
\label{appendix_a}
GPT4Vis \cite{wu2023gpt4vis} centers on the evaluation of GPT4's linguistic and visual capabilities in zero-shot visual recognition tasks. Specifically, it explores the potential of GPT4's generated rich textual descriptions of categories to enhance recognition performance without any training across 16 images, videos and point clouds datasets. For the video recognition datasets, an example of the instruction and GPT4's response is as follows:
\begin{quote}
\textbf{\textit{Q:}} \textit{I want you to act as a human action expert. I will give you a word and your task is to give me $K$ sentences to describe the word. Your description must accurately revolve around this word and be as objective, detailed and diverse as possible. In addition, the subject of your description is a human action video. Do not provide any explanations. The word is ``climbing a rope".}
\end{quote}
\begin{quote}
\textbf{\textit{A:}} \textit{``Gripping the rope tightly, they began the ascent with determination etched on their face.",
        ``Wrapping their legs around the rope, they pushed upward using the strength of their thighs.",
        ``They reached for the next section of rope, pulling themselves higher with steady breaths.",
        ``Muscles flexed and tensed as each hand-over-hand movement brought them closer to the top.",
        ``A bead of sweat trickled down their forehead, a testament to the effort of climbing the rope.",
        ``Exhaling sharply, they conquered another segment of the vertical challenge, not looking down.",
        ``Their arms ached from the exertion, yet they persisted in climbing the rope with unwavering resolve.",
        ``The rough texture of the rope grazed their palms, each grasp bringing a slight sting.",
        ``They paused momentarily to catch their breath, hanging from the rope with practiced ease.",
        ``With each upward motion, they felt the burn in their shoulders intensify, driving them forward.}
\end{quote}
It generates $K$ descriptive sentences for each category, with $K$ defaulting to be 20. In our paper, we sampled 10 sentences for each category as its Generic prompt.

\section{More examples of Discriminative prompt}
\label{appendix_b}





Using \textit{climbing a rope} as an example, its top-5 semantically similar neighbor is \textit{\{rock climbing, climbing ladder, abseiling, climbing tree, swinging on something\}}, and the Discriminative prompt is as follows:

\begin{quote}
\textit{``A person tightly grips a vertically suspended rope with both hands and feet, using coordinated pulling and pushing movements to ascend the rope steadily, without interacting with any rigid structures like rocks, ladders, or trees."}
\end{quote}

Using \textit{diving cliff} as an example, its top-5 semantically similar neighbor is \textit{\{springboard diving, scuba diving, rock climbing, skydiving, surfing water\}}, and the Discriminative prompt is as follows:

\begin{quote}
\textit{``A video showcasing cliff diving would depict an individual leaping from a high rock or natural formation into a body of water, focusing on the aerial descent, the body posture during freefall, and the splash upon entry, with no use of diving boards, equipment, or parachutes."}
\end{quote}

Using \textit{yoga} as an example, its top-5 semantically similar neighbor is \textit{\{tai chi, stretching leg, bending back, doing aerobics, dancing ballet\}}, and the Discriminative prompt is as follows:

\begin{quote}
\textit{``A person holds a balanced, stationary pose such as a downward-facing dog or tree pose, with deliberate breathing patterns, slow and controlled transitions between postures, and a calm, focused demeanor, avoiding rapid or rhythmic movements."}
\end{quote}

\section{Examples of frame and video descriptions with temporal saliency in KGTA}
\label{appendix_c}

\begin{figure*}[!h]
\vspace{-0.1cm}
  \centering
  \includegraphics[width=1.0\textwidth]{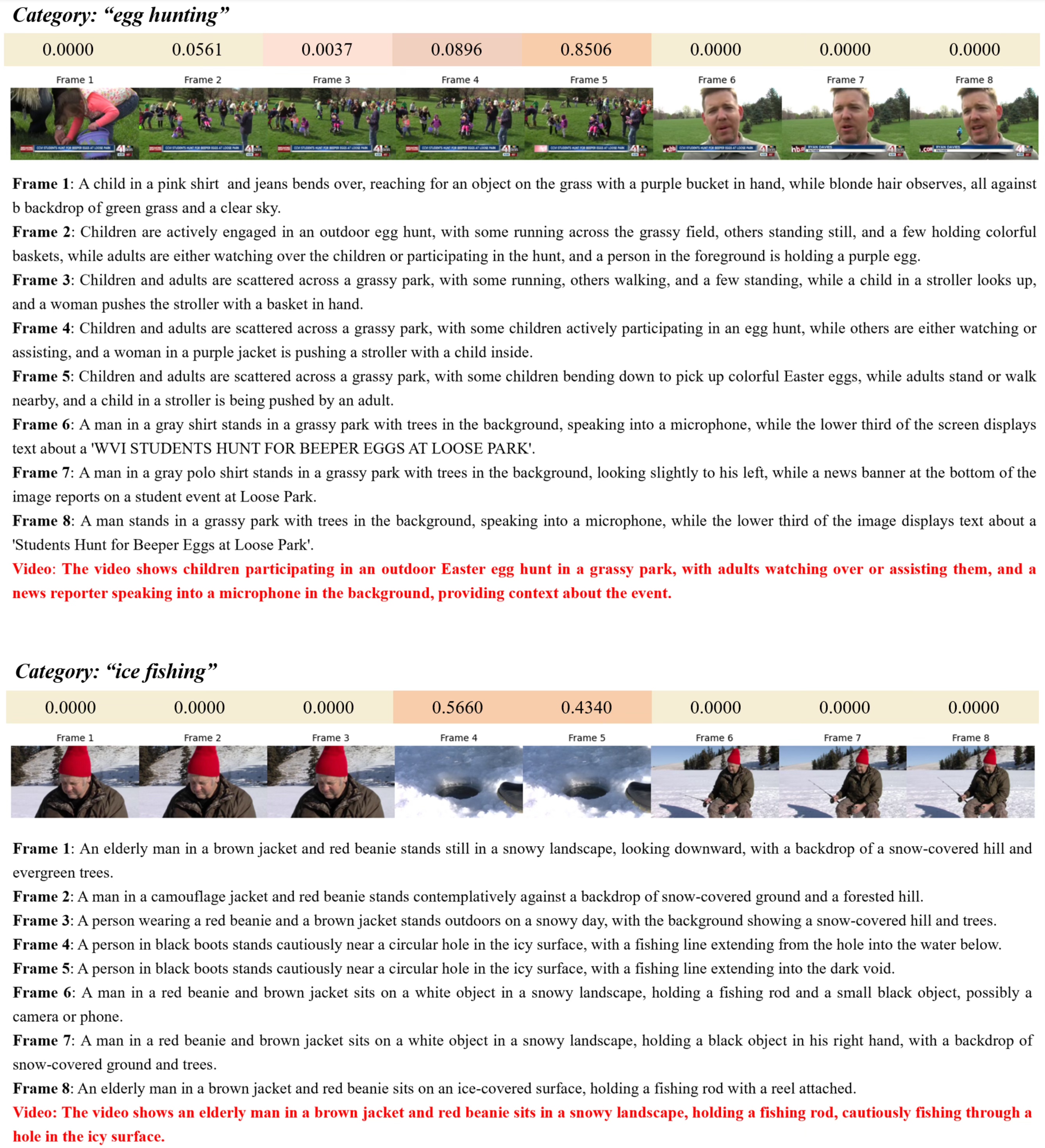}
  \caption{Examples of frame descriptions and video descriptions, along with the temporal saliencies.}
  \label{figs1}
\end{figure*}

We present some examples of the CogVLM2-generated frame descriptions and Llama-3.1-generated video descriptions in Fig. \ref{figs1}, where the temporal saliency calculated from the similarities between the frame descriptions and video description in KGTA is labeled above the frame sequence.


\section{Details of datasets and evaluation protocols in zero-shot setting}
\label{appendix_d}

\subsection{Details of datasets}
We describe the video datasets used in our experiments:


%
\begin{itemize}
    \item \noindent\textbf{Kinetics-400} is a large scale video dataset that includes 240,000 training videos and 20,000 validation videos across 400 different action categories from YouTube. Each video clip is roughly 10 seconds in length, capturing a concise action moment.
    \item \noindent\textbf{Kinetics-600} builds upon the foundation set by Kinetics-400, introducing an additional 220 action categories that enrich the dataset, particularly for evaluating zero-shot learning capabilities.
    \item \noindent\textbf{UCF-101} is an action recognition dataset that contains 13,320 videos from 101 realistic action categories, collected from YouTube. Each video clip is a succinct representation of an action, averaging 7.21 seconds in length.
    \item \noindent\textbf{HMDB-51} is a video recognition dataset that contains 7,000 video clips from 51 action categories. It is a collection of realistic videos from various sources, including movies and web videos. 
\end{itemize}

\subsection{Evaluation protocols in zero-shot setting}

We employ the Kinetics-400 pre-trained model to evaluate on UCF-101, HMDB-51 and Kinetics-600 datasets, following the evaluation protocols that are widely used.

\noindent\textbf{UCF-101$\&$HMDB-51:} 

\begin{itemize}
    \item \noindent\textbf{Evaluation Protocol 1 (EP1):} We first randomly choose half of the classes in the test dataset, \ie, 50 for UCF-101 and 25 for HMDB-51, and repeat ten times and report averaged result and standard deviation for each datasets. \cite{brattoli2020rethinking}
    \item \noindent\textbf{Evaluation Protocol 2 (EP2):} We test the pre-trained model on full UCF-101 and HMAB-51, \ie, evaluating on all 101 UCF classes and all 51 HMDB classes. \cite{brattoli2020rethinking}
    \item \noindent\textbf{Evaluation Protocol 3 (EP3):} We perform testing using three oficial splits and average the results of each split. The average top-1 accuracy and standard deviation are reported. \cite{ni2022expanding}
\end{itemize}

\noindent\textbf{Kinetics-600:} 
\begin{itemize}
    \item \cite{chen2021elaborative} randomly divided the 220 new classes in Kinetics-600 into 60 validation classes and 160 testing classes, repeating this process three times. We conduct test on the three splits provided in \cite{chen2021elaborative} and report the average top-1 accuracy and the standard deviation.
\end{itemize}

\section{More training details}
\label{appendix_e}

In zero-shot setting, we use 2 epochs to warm up the training and another 20 epochs for fine-tuning on Kinetics-400. In few-shot setting, we use 2 epochs to warm up the training and another 8 epochs for fine-tuning on the target datasets, with no pre-training on Kinetics-400. For both zero-shot and few-shot settings, the learning rate for warm up is $3.33 \times {10^{ - 8}}$. The initial learning rate is $3.33 \times {10^{ - 6}}$ and is decayed to $3.33 \times {10^{ - 8}}$ following a cosine schedule. We use ViT-B/16 based CLIP model for all experiments and extract 8 frames for each video and the spatial resolution of the input frame is $224 \times 224$. We use the pre-trained CogVLM2-llama3-chat-19B-int4 version for generating frame descriptions, and use Llama-3.1-8B-Instruct version for aggregating them into video description. The experiments are conducted on 8$\times$A6000GPUs. Training the model requires almost a day.

\section{More visualizations of inter-class correlation maps}
\label{appendix_f}
\begin{figure*}[!htp]

\vspace{+3cm}

    \centering

    \begin{minipage}[b]{0.38\textwidth}
        \centering
        \includegraphics[width=\linewidth]{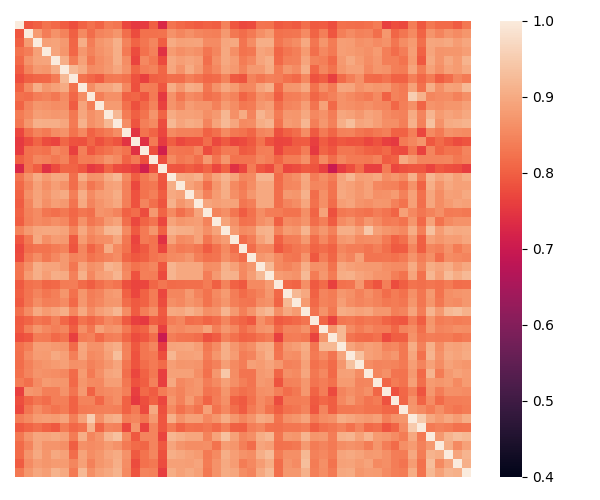}
        \caption{Inter-class correlation map of 51 categories of HMDB-51 with Hand-crafted prompt.}\label{figs2}
    \end{minipage}%
    \hspace{0.7cm}
    \begin{minipage}[b]{0.38\textwidth}
        \centering
        \includegraphics[width=\linewidth]{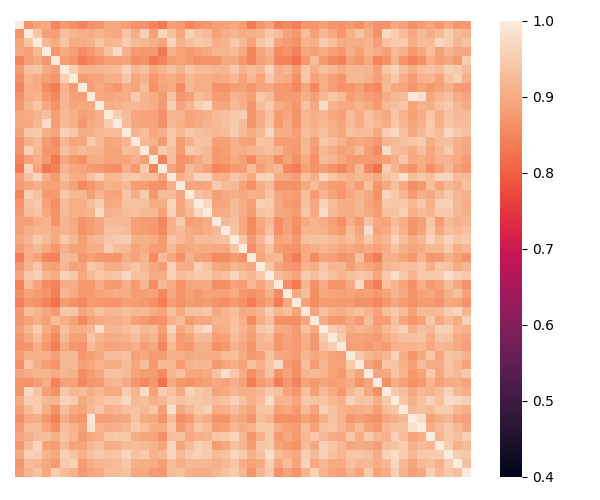}
        \caption{Inter-class correlation map of 51 categories of HMDB-51 with Generic prompt.}\label{figs3}
    \end{minipage}
    \begin{minipage}[b]{0.38\textwidth}
        \centering
        \includegraphics[width=\linewidth]{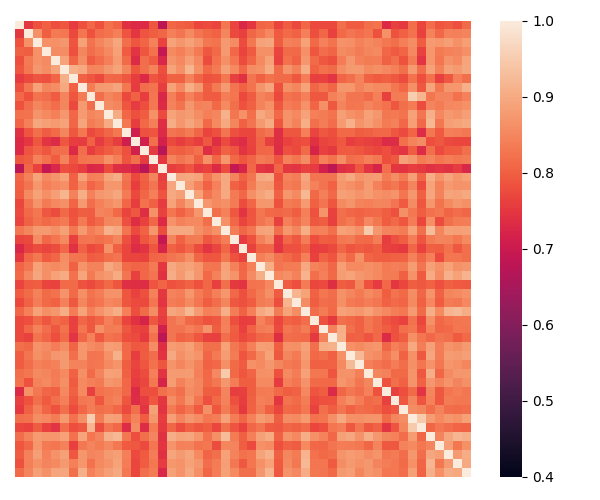}
        \caption{Inter-class correlation map of 51 categories of HMDB-51 with Discriminative prompt.}\label{figs4}
    \end{minipage}%
    \hspace{0.7cm}
    \begin{minipage}[b]{0.38\textwidth}
        \centering
        \includegraphics[width=\linewidth]{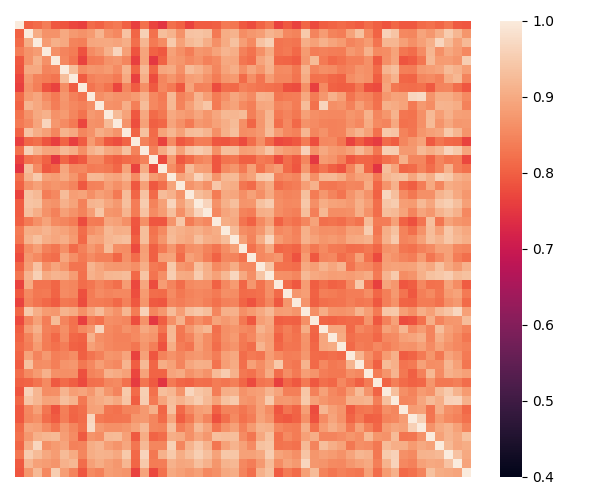}
        \caption{Inter-class correlation map of 51 categories of HMDB-51 with GD-prompt.}\label{figs5}
    \end{minipage}
\vspace{+3cm}

\end{figure*}

\begin{figure*}[h]
\vspace{+4cm}
    \centering

    \begin{minipage}[b]{0.38\textwidth}
        \centering
        \includegraphics[width=\linewidth]{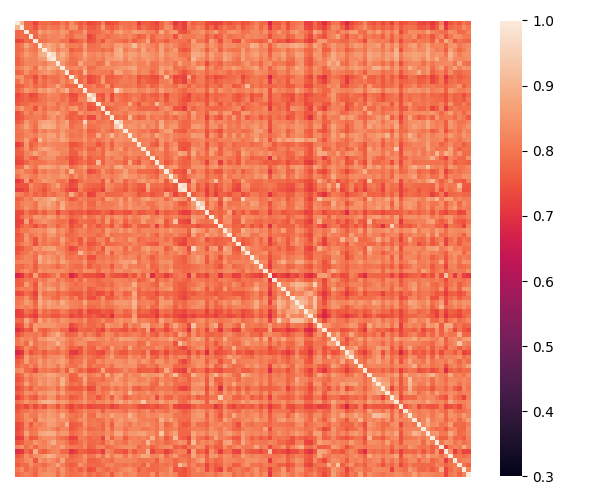}
        \caption{Inter-class correlation map of 101 categories of UCF-101 with Hand-crafted prompt.}\label{figs6}
    \end{minipage}%
    \hspace{0.7cm}
    \begin{minipage}[b]{0.38\textwidth}
        \centering
        \includegraphics[width=\linewidth]{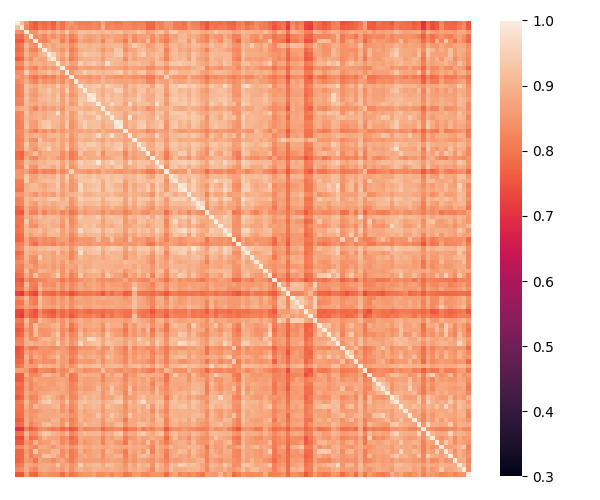}
        \caption{Inter-class correlation map of 101 categories of UCF-101 with gENERIC prompt.}\label{figs7}
    \end{minipage}
    
    \begin{minipage}[b]{0.38\textwidth}
        \centering
        \includegraphics[width=\linewidth]{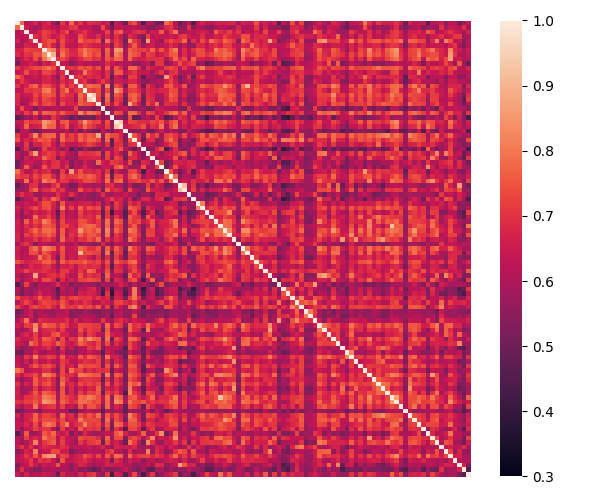}
        \caption{Inter-class correlation map of 101 categories of UCF-101 with Discriminative prompt.}\label{figs8}
    \end{minipage}%
    \hspace{0.7cm}
    \begin{minipage}[b]{0.38\textwidth}
        \centering
        \includegraphics[width=\linewidth]{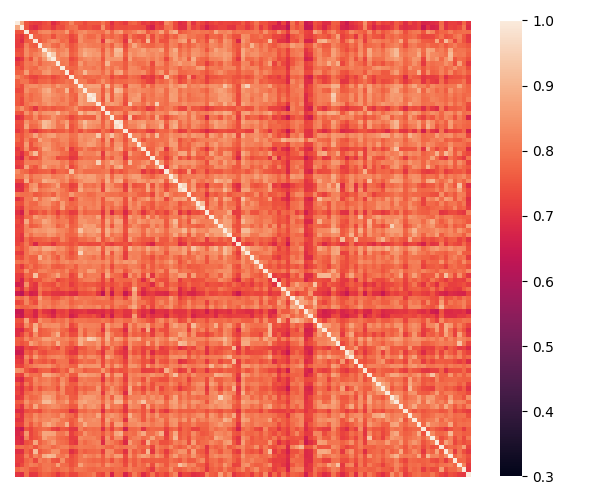}
        \caption{Inter-class correlation map of 101 categories of UCF-101 with GD-prompt.}\label{figs9}
    \end{minipage}
    \vspace{+5cm}
\end{figure*}

We provide more visualizations of inter-class correlation maps of UCF-101 and HMDB-51 with different prompts in Fig. \ref{figs2}--\ref{figs9}. The darker the color, the lower the similarity.


%% file: checklist.tex
\section*{NeurIPS Paper Checklist}

\begin{enumerate}

\item {\bf Claims}
    \item[] Question: Do the main claims made in the abstract and introduction accurately reflect the paper's contributions and scope?
    \item[] Answer: \answerYes{} 
    \item[] Justification: See abstract and Sec. \ref{sec:intro}

\item {\bf Limitations}
    \item[] Question: Does the paper discuss the limitations of the work performed by the authors?
    \item[] Answer: \answerYes{} 
    \item[] Justification: See Sec. \ref{sec:conclu}

\item {\bf Theory assumptions and proofs}
    \item[] Question: For each theoretical result, does the paper provide the full set of assumptions and a complete (and correct) proof?
    \item[] Answer: \answerNA{} 
    \item[] Justification: This paper does not include theoretical results.

    \item {\bf Experimental result reproducibility}
    \item[] Question: Does the paper fully disclose all the information needed to reproduce the main experimental results of the paper to the extent that it affects the main claims and/or conclusions of the paper (regardless of whether the code and data are provided or not)?
    \item[] Answer: \answerYes{} 
    \item[] Justification: We describe the experiment settings and training details in Sec. \ref{sec:4.1} and Appendix \ref{appendix_e}

\item {\bf Open access to data and code}
    \item[] Question: Does the paper provide open access to the data and code, with sufficient instructions to faithfully reproduce the main experimental results, as described in supplemental material?
    \item[] Answer: \answerYes{} 
    \item[] Justification: Code will be released after blind review.

\item {\bf Experimental setting/details}
    \item[] Question: Does the paper specify all the training and test details (e.g., data splits, hyperparameters, how they were chosen, type of optimizer, etc.) necessary to understand the results?
    \item[] Answer: \answerYes{} 
    \item[] Justification: See experiment settings and training details in Sec. \ref{sec:4.1} and Appendix \ref{appendix_e}

\item {\bf Experiment statistical significance}
    \item[] Question: Does the paper report error bars suitably and correctly defined or other appropriate information about the statistical significance of the experiments?
    \item[] Answer: \answerYes{} 
    \item[] Justification: Under zero-shot setting, we evaluate the model on UCF-101, HMDB-51 and K-600 using different test data splits, and report the average Top-1 accuracy and standard deviation. See Appendix \ref{appendix_d}.

\item {\bf Experiments compute resources}
    \item[] Question: For each experiment, does the paper provide sufficient information on the computer resources (type of compute workers, memory, time of execution) needed to reproduce the experiments?
    \item[] Answer: \answerYes{} 
    \item[] Justification: See Appendix \ref{appendix_e}
    
\item {\bf Code of ethics}
    \item[] Question: Does the research conducted in the paper conform, in every respect, with the NeurIPS Code of Ethics \url{https://neurips.cc/public/EthicsGuidelines}?
    \item[] Answer: \answerYes{} 
    \item[] Justification: The research in this paper conforms the NeurIPS Code of Ethics.

\item {\bf Broader impacts}
    \item[] Question: Does the paper discuss both potential positive societal impacts and negative societal impacts of the work performed?
    \item[] Answer: \answerNA{} 
    \item[] Justification: This paper focus on a classification task, with no privacy considerations, and security considerations.
    
\item {\bf Safeguards}
    \item[] Question: Does the paper describe safeguards that have been put in place for responsible release of data or models that have a high risk for misuse (e.g., pretrained language models, image generators, or scraped datasets)?
    \item[] Answer: \answerNA{} 
    \item[] Justification: This paper poses no such risks.

\item {\bf Licenses for existing assets}
    \item[] Question: Are the creators or original owners of assets (e.g., code, data, models), used in the paper, properly credited and are the license and terms of use explicitly mentioned and properly respected?
    \item[] Answer: \answerYes{} 
    \item[] Justification: All data and models are mentioned.

\item {\bf New assets}
    \item[] Question: Are new assets introduced in the paper well documented and is the documentation provided alongside the assets?
    \item[] Answer: \answerNo{} 
    \item[] Justification: This paper does not release new assets.

\item {\bf Crowdsourcing and research with human subjects}
    \item[] Question: For crowdsourcing experiments and research with human subjects, does the paper include the full text of instructions given to participants and screenshots, if applicable, as well as details about compensation (if any)? 
    \item[] Answer: \answerNo{} 
    \item[] Justification: This paper does not involve crowdsourcing nor research with human subjects.

\item {\bf Institutional review board (IRB) approvals or equivalent for research with human subjects}
    \item[] Question: Does the paper describe potential risks incurred by study participants, whether such risks were disclosed to the subjects, and whether Institutional Review Board (IRB) approvals (or an equivalent approval/review based on the requirements of your country or institution) were obtained?
    \item[] Answer: \answerNA{} 
    \item[] Justification: This paper does not involve crowdsourcing nor research with human subjects.

\item {\bf Declaration of LLM usage}
    \item[] Question: Does the paper describe the usage of LLMs if it is an important, original, or non-standard component of the core methods in this research? Note that if the LLM is used only for writing, editing, or formatting purposes and does not impact the core methodology, scientific rigorousness, or originality of the research, declaration is not required.
    \item[] Answer: \answerYes{} 
    \item[] LLMs are used as an important component of this research to generate natural language prompts.

\end{enumerate}

%% file: egbib.bib
@inproceedings{radford2021learning,
  title={Learning transferable visual models from natural language supervision},
  author={Radford, Alec and Kim, Jong Wook and Hallacy, Chris and Ramesh, Aditya and Goh, Gabriel and Agarwal, Sandhini and Sastry, Girish and Askell, Amanda and Mishkin, Pamela and Clark, Jack and others},
  booktitle={International conference on machine learning},
  pages={8748--8763},
  year={2021},
  organization={PMLR}
}

@inproceedings{jia2021scaling,
  title={Scaling up visual and vision-language representation learning with noisy text supervision},
  author={Jia, Chao and Yang, Yinfei and Xia, Ye and Chen, Yi-Ting and Parekh, Zarana and Pham, Hieu and Le, Quoc and Sung, Yun-Hsuan and Li, Zhen and Duerig, Tom},
  booktitle={International conference on machine learning},
  pages={4904--4916},
  year={2021},
  organization={PMLR}
}

@article{yuan2021florence,
  title={Florence: A new foundation model for computer vision},
  author={Yuan, Lu and Chen, Dongdong and Chen, Yi-Ling and Codella, Noel and Dai, Xiyang and Gao, Jianfeng and Hu, Houdong and Huang, Xuedong and Li, Boxin and Li, Chunyuan and others},
  journal={arXiv preprint arXiv:2111.11432},
  year={2021}
}

@inproceedings{wu2023bidirectional,
  title={Bidirectional cross-modal knowledge exploration for video recognition with pre-trained vision-language models},
  author={Wu, Wenhao and Wang, Xiaohan and Luo, Haipeng and Wang, Jingdong and Yang, Yi and Ouyang, Wanli},
  booktitle={Proceedings of the IEEE/CVF conference on computer vision and pattern recognition},
  pages={6620--6630},
  year={2023}
}

@article{wang2021actionclip,
  title={Actionclip: A new paradigm for video action recognition},
  author={Wang, Mengmeng and Xing, Jiazheng and Liu, Yong},
  journal={arXiv preprint arXiv:2109.08472},
  year={2021}
}

@inproceedings{ni2022expanding,
  title={Expanding language-image pretrained models for general video recognition},
  author={Ni, Bolin and Peng, Houwen and Chen, Minghao and Zhang, Songyang and Meng, Gaofeng and Fu, Jianlong and Xiang, Shiming and Ling, Haibin},
  booktitle={European Conference on Computer Vision},
  pages={1--18},
  year={2022},
  organization={Springer}
}

@article{wang2024m2,
  title={M2-CLIP: A Multimodal, Multi-task Adapting Framework for Video Action Recognition},
  author={Wang, Mengmeng and Xing, Jiazheng and Jiang, Boyuan and Chen, Jun and Mei, Jianbiao and Zuo, Xingxing and Dai, Guang and Wang, Jingdong and Liu, Yong},
  journal={arXiv preprint arXiv:2401.11649},
  year={2024}
}

@article{huang2024froster,
  title={Froster: Frozen clip is a strong teacher for open-vocabulary action recognition},
  author={Huang, Xiaohu and Zhou, Hao and Yao, Kun and Han, Kai},
  journal={arXiv preprint arXiv:2402.03241},
  year={2024}
}

@article{wu2024building,
  title={Building an open-vocabulary video CLIP model with better architectures, optimization and data},
  author={Wu, Zuxuan and Weng, Zejia and Peng, Wujian and Yang, Xitong and Li, Ang and Davis, Larry S and Jiang, Yu-Gang},
  journal={IEEE Transactions on Pattern Analysis and Machine Intelligence},
  year={2024},
  publisher={IEEE}
}

@inproceedings{li2023blip,
  title={Blip-2: Bootstrapping language-image pre-training with frozen image encoders and large language models},
  author={Li, Junnan and Li, Dongxu and Savarese, Silvio and Hoi, Steven},
  booktitle={International conference on machine learning},
  pages={19730--19742},
  year={2023},
  organization={PMLR}
}

@article{touvron2023llama,
  title={Llama 2: Open foundation and fine-tuned chat models},
  author={Touvron, Hugo and Martin, Louis and Stone, Kevin and Albert, Peter and Almahairi, Amjad and Babaei, Yasmine and Bashlykov, Nikolay and Batra, Soumya and Bhargava, Prajjwal and Bhosale, Shruti and others},
  journal={arXiv preprint arXiv:2307.09288},
  year={2023}
}

@inproceedings{carreira2017quo,
  title={Quo vadis, action recognition? a new model and the kinetics dataset},
  author={Carreira, Joao and Zisserman, Andrew},
  booktitle={proceedings of the IEEE Conference on Computer Vision and Pattern Recognition},
  pages={6299--6308},
  year={2017}
}

@article{soomro2012ucf101,
  title={UCF101: A dataset of 101 human actions classes from videos in the wild},
  author={Soomro, K},
  journal={arXiv preprint arXiv:1212.0402},
  year={2012}
}

@article{wu2023gpt4vis,
  title={GPT4Vis: what can GPT-4 do for zero-shot visual recognition?},
  author={Wu, Wenhao and Yao, Huanjin and Zhang, Mengxi and Song, Yuxin and Ouyang, Wanli and Wang, Jingdong},
  journal={arXiv preprint arXiv:2311.15732},
  year={2023}
}

@inproceedings{wu2023revisiting,
  title={Revisiting classifier: Transferring vision-language models for video recognition},
  author={Wu, Wenhao and Sun, Zhun and Ouyang, Wanli},
  booktitle={Proceedings of the AAAI conference on artificial intelligence},
  volume={37},
  number={3},
  pages={2847--2855},
  year={2023}
}

@inproceedings{tu2023implicit,
  title={Implicit temporal modeling with learnable alignment for video recognition},
  author={Tu, Shuyuan and Dai, Qi and Wu, Zuxuan and Cheng, Zhi-Qi and Hu, Han and Jiang, Yu-Gang},
  booktitle={Proceedings of the IEEE/CVF International Conference on Computer Vision},
  pages={19936--19947},
  year={2023}
}

@inproceedings{ju2022prompting,
  title={Prompting visual-language models for efficient video understanding},
  author={Ju, Chen and Han, Tengda and Zheng, Kunhao and Zhang, Ya and Xie, Weidi},
  booktitle={European Conference on Computer Vision},
  pages={105--124},
  year={2022},
  organization={Springer}
}

@article{pan2022st,
  title={St-adapter: Parameter-efficient image-to-video transfer learning},
  author={Pan, Junting and Lin, Ziyi and Zhu, Xiatian and Shao, Jing and Li, Hongsheng},
  journal={Advances in Neural Information Processing Systems},
  volume={35},
  pages={26462--26477},
  year={2022}
}

@article{yang2023aim,
  title={Aim: Adapting image models for efficient video action recognition},
  author={Yang, Taojiannan and Zhu, Yi and Xie, Yusheng and Zhang, Aston and Chen, Chen and Li, Mu},
  journal={arXiv preprint arXiv:2302.03024},
  year={2023}
}

@inproceedings{lin2022frozen,
  title={Frozen clip models are efficient video learners},
  author={Lin, Ziyi and Geng, Shijie and Zhang, Renrui and Gao, Peng and De Melo, Gerard and Wang, Xiaogang and Dai, Jifeng and Qiao, Yu and Li, Hongsheng},
  booktitle={European Conference on Computer Vision},
  pages={388--404},
  year={2022},
  organization={Springer}
}

@inproceedings{liu2023revisiting,
  title={Revisiting temporal modeling for clip-based image-to-video knowledge transferring},
  author={Liu, Ruyang and Huang, Jingjia and Li, Ge and Feng, Jiashi and Wu, Xinglong and Li, Thomas H},
  booktitle={Proceedings of the IEEE/CVF Conference on Computer Vision and Pattern Recognition},
  pages={6555--6564},
  year={2023}
}

@inproceedings{wasim2023vita,
  title={Vita-clip: Video and text adaptive clip via multimodal prompting},
  author={Wasim, Syed Talal and Naseer, Muzammal and Khan, Salman and Khan, Fahad Shahbaz and Shah, Mubarak},
  booktitle={Proceedings of the IEEE/CVF Conference on Computer Vision and Pattern Recognition},
  pages={23034--23044},
  year={2023}
}

@inproceedings{park2023dual,
  title={Dual-path adaptation from image to video transformers},
  author={Park, Jungin and Lee, Jiyoung and Sohn, Kwanghoon},
  booktitle={Proceedings of the IEEE/CVF Conference on Computer Vision and Pattern Recognition},
  pages={2203--2213},
  year={2023}
}

@inproceedings{weng2023open,
  title={Open-vclip: Transforming clip to an open-vocabulary video model via interpolated weight optimization},
  author={Weng, Zejia and Yang, Xitong and Li, Ang and Wu, Zuxuan and Jiang, Yu-Gang},
  booktitle={International Conference on Machine Learning},
  pages={36978--36989},
  year={2023},
  organization={PMLR}
}

@inproceedings{jia2024generating,
  title={Generating action-conditioned prompts for open-vocabulary video action recognition},
  author={Jia, Chengyou and Luo, Minnan and Chang, Xiaojun and Dang, Zhuohang and Han, Mingfei and Wang, Mengmeng and Dai, Guang and Dang, Sizhe and Wang, Jingdong},
  booktitle={Proceedings of the 32nd ACM International Conference on Multimedia},
  pages={4640--4649},
  year={2024}
}

@article{zhou2022learning,
  title={Learning to prompt for vision-language models},
  author={Zhou, Kaiyang and Yang, Jingkang and Loy, Chen Change and Liu, Ziwei},
  journal={International Journal of Computer Vision},
  volume={130},
  number={9},
  pages={2337--2348},
  year={2022},
  publisher={Springer}
}

@inproceedings{zhou2022conditional,
  title={Conditional prompt learning for vision-language models},
  author={Zhou, Kaiyang and Yang, Jingkang and Loy, Chen Change and Liu, Ziwei},
  booktitle={Proceedings of the IEEE/CVF conference on computer vision and pattern recognition},
  pages={16816--16825},
  year={2022}
}

@inproceedings{jia2022visual,
  title={Visual prompt tuning},
  author={Jia, Menglin and Tang, Luming and Chen, Bor-Chun and Cardie, Claire and Belongie, Serge and Hariharan, Bharath and Lim, Ser-Nam},
  booktitle={European Conference on Computer Vision},
  pages={709--727},
  year={2022},
  organization={Springer}
}

@inproceedings{khattak2023maple,
  title={Maple: Multi-modal prompt learning},
  author={Khattak, Muhammad Uzair and Rasheed, Hanoona and Maaz, Muhammad and Khan, Salman and Khan, Fahad Shahbaz},
  booktitle={Proceedings of the IEEE/CVF Conference on Computer Vision and Pattern Recognition},
  pages={19113--19122},
  year={2023}
}

@article{zang2022unified,
  title={Unified vision and language prompt learning},
  author={Zang, Yuhang and Li, Wei and Zhou, Kaiyang and Huang, Chen and Loy, Chen Change},
  journal={arXiv preprint arXiv:2210.07225},
  year={2022}
}

@inproceedings{goswami2024copl,
  title={CoPL: Contextual Prompt Learning for Vision-Language Understanding},
  author={Goswami, Koustava and Karanam, Srikrishna and Udhayanan, Prateksha and Joseph, KJ and Srinivasan, Balaji Vasan},
  booktitle={Proceedings of the AAAI Conference on Artificial Intelligence},
  volume={38},
  number={16},
  pages={18090--18098},
  year={2024}
}

@article{yu2022coca,
  title={Coca: Contrastive captioners are image-text foundation models},
  author={Yu, Jiahui and Wang, Zirui and Vasudevan, Vijay and Yeung, Legg and Seyedhosseini, Mojtaba and Wu, Yonghui},
  journal={arXiv preprint arXiv:2205.01917},
  year={2022}
}

@inproceedings{qing2023disentangling,
  title={Disentangling spatial and temporal learning for efficient image-to-video transfer learning},
  author={Qing, Zhiwu and Zhang, Shiwei and Huang, Ziyuan and Zhang, Yingya and Gao, Changxin and Zhao, Deli and Sang, Nong},
  booktitle={Proceedings of the IEEE/CVF International Conference on Computer Vision},
  pages={13934--13944},
  year={2023}
}

@article{bahng2022exploring,
  title={Exploring visual prompts for adapting large-scale models},
  author={Bahng, Hyojin and Jahanian, Ali and Sankaranarayanan, Swami and Isola, Phillip},
  journal={arXiv preprint arXiv:2203.17274},
  year={2022}
}

@inproceedings{yao2023visual,
  title={Visual-language prompt tuning with knowledge-guided context optimization},
  author={Yao, Hantao and Zhang, Rui and Xu, Changsheng},
  booktitle={Proceedings of the IEEE/CVF conference on computer vision and pattern recognition},
  pages={6757--6767},
  year={2023}
}

@inproceedings{zhu2023prompt,
  title={Prompt-aligned gradient for prompt tuning},
  author={Zhu, Beier and Niu, Yulei and Han, Yucheng and Wu, Yue and Zhang, Hanwang},
  booktitle={Proceedings of the IEEE/CVF International Conference on Computer Vision},
  pages={15659--15669},
  year={2023}
}

@inproceedings{bulat2023lasp,
  title={Lasp: Text-to-text optimization for language-aware soft prompting of vision \& language models},
  author={Bulat, Adrian and Tzimiropoulos, Georgios},
  booktitle={Proceedings of the IEEE/CVF Conference on Computer Vision and Pattern Recognition},
  pages={23232--23241},
  year={2023}
}

@inproceedings{maniparambil2023enhancing,
  title={Enhancing clip with gpt-4: Harnessing visual descriptions as prompts},
  author={Maniparambil, Mayug and Vorster, Chris and Molloy, Derek and Murphy, Noel and McGuinness, Kevin and O'Connor, Noel E},
  booktitle={Proceedings of the IEEE/CVF International Conference on Computer Vision},
  pages={262--271},
  year={2023}
}

@article{menon2022visual,
  title={Visual classification via description from large language models},
  author={Menon, Sachit and Vondrick, Carl},
  journal={arXiv preprint arXiv:2210.07183},
  year={2022}
}

@inproceedings{pratt2023does,
  title={What does a platypus look like? generating customized prompts for zero-shot image classification},
  author={Pratt, Sarah and Covert, Ian and Liu, Rosanne and Farhadi, Ali},
  booktitle={Proceedings of the IEEE/CVF International Conference on Computer Vision},
  pages={15691--15701},
  year={2023}
}

@inproceedings{yang2023language,
  title={Language in a bottle: Language model guided concept bottlenecks for interpretable image classification},
  author={Yang, Yue and Panagopoulou, Artemis and Zhou, Shenghao and Jin, Daniel and Callison-Burch, Chris and Yatskar, Mark},
  booktitle={Proceedings of the IEEE/CVF Conference on Computer Vision and Pattern Recognition},
  pages={19187--19197},
  year={2023}
}

@article{su2022language,
  title={Language models can see: Plugging visual controls in text generation},
  author={Su, Yixuan and Lan, Tian and Liu, Yahui and Liu, Fangyu and Yogatama, Dani and Wang, Yan and Kong, Lingpeng and Collier, Nigel},
  journal={arXiv preprint arXiv:2205.02655},
  year={2022}
}

@inproceedings{yang2022empirical,
  title={An empirical study of gpt-3 for few-shot knowledge-based vqa},
  author={Yang, Zhengyuan and Gan, Zhe and Wang, Jianfeng and Hu, Xiaowei and Lu, Yumao and Liu, Zicheng and Wang, Lijuan},
  booktitle={Proceedings of the AAAI conference on artificial intelligence},
  volume={36},
  number={3},
  pages={3081--3089},
  year={2022}
}

@article{wang2023cogvlm,
  title={Cogvlm: Visual expert for pretrained language models},
  author={Wang, Weihan and Lv, Qingsong and Yu, Wenmeng and Hong, Wenyi and Qi, Ji and Wang, Yan and Ji, Junhui and Yang, Zhuoyi and Zhao, Lei and Song, Xixuan and others},
  journal={arXiv preprint arXiv:2311.03079},
  year={2023}
}

@article{dubey2024llama,
  title={The llama 3 herd of models},
  author={Dubey, Abhimanyu and Jauhri, Abhinav and Pandey, Abhinav and Kadian, Abhishek and Al-Dahle, Ahmad and Letman, Aiesha and Mathur, Akhil and Schelten, Alan and Yang, Amy and Fan, Angela and others},
  journal={arXiv preprint arXiv:2407.21783},
  year={2024}
}

@inproceedings{kuehne2011hmdb,
  title={HMDB: a large video database for human motion recognition},
  author={Kuehne, Hildegard and Jhuang, Hueihan and Garrote, Est{\'\i}baliz and Poggio, Tomaso and Serre, Thomas},
  booktitle={2011 International conference on computer vision},
  pages={2556--2563},
  year={2011},
  organization={IEEE}
}

@inproceedings{brattoli2020rethinking,
  title={Rethinking zero-shot video classification: End-to-end training for realistic applications},
  author={Brattoli, Biagio and Tighe, Joseph and Zhdanov, Fedor and Perona, Pietro and Chalupka, Krzysztof},
  booktitle={Proceedings of the IEEE/CVF conference on computer vision and pattern recognition},
  pages={4613--4623},
  year={2020}
}

@inproceedings{chen2021elaborative,
  title={Elaborative rehearsal for zero-shot action recognition},
  author={Chen, Shizhe and Huang, Dong},
  booktitle={Proceedings of the IEEE/CVF international conference on computer vision},
  pages={13638--13647},
  year={2021}
}

@inproceedings{rasheed2023fine,
  title={Fine-tuned clip models are efficient video learners},
  author={Rasheed, Hanoona and Khattak, Muhammad Uzair and Maaz, Muhammad and Khan, Salman and Khan, Fahad Shahbaz},
  booktitle={Proceedings of the IEEE/CVF conference on computer vision and pattern recognition},
  pages={6545--6554},
  year={2023}
}

@article{carreira2018short,
  title={A short note about kinetics-600},
  author={Carreira, Joao and Noland, Eric and Banki-Horvath, Andras and Hillier, Chloe and Zisserman, Andrew},
  journal={arXiv preprint arXiv:1808.01340},
  year={2018}
}
